\pdfoutput=1
\documentclass[conference]{IEEEtran}

\usepackage[T1]{fontenc}
\usepackage{amsmath}
\usepackage{amssymb}
\usepackage{array}
\usepackage{booktabs}
\usepackage{threeparttable}
\usepackage{graphicx}
\usepackage{url}
\usepackage[hidelinks]{hyperref}
\usepackage{xcolor}
\usepackage{tikz}
\usetikzlibrary{positioning,arrows.meta,fit,calc}

\newcommand{\fieldname}{DetField}
\newcommand{\papertitle}{An Image Detectability Field as a Viewpoint Utility for 2D--3D Localization in Sparse Maps}

\newcommand{\meth}[1]{\textsf{#1}}
\newcommand{\arm}[1]{\textsf{#1}}

\DeclareMathOperator*{\argmax}{arg\,max}

\begin{document}
\bstctlcite{IEEEexample:BSTcontrol}

\title{\papertitle}
\author{\IEEEauthorblockN{Han Sun\IEEEauthorrefmark{1}, Jun Li and Christy Jie Liang}
\IEEEauthorblockA{Faculty of Engineering and Information Technology,
University of Technology Sydney, Ultimo NSW 2007, Australia\\
\texttt{\{han.sun@student, jun.li@, jie.liang@\}.uts.edu.au}}}

\maketitle
\begingroup\renewcommand\thefootnote{\IEEEauthorrefmark{1}}
\footnotetext{Corresponding author.}
\endgroup

\begin{abstract}
In 2D--3D localization, a camera estimates its pose by matching an image
to a 3D map. A camera held in a hand or mounted on a robot can often choose
its viewpoint before it takes the image. In this work, we ask whether an image detectability field is a useful
viewpoint utility when the planner reads only the geometry of the map. The
field can be queried at any 3D point near a surface. From the local geometry
around the point, it predicts how likely a 2D image detector is to fire at
the projection of the point. We use the trained field as it is, without
fine-tuning it for view selection. A fixed SuperPoint and EPnP-RANSAC
localizer then estimates the pose from the chosen view. We answer in two settings, each with a test registered before the runs.
On synthetic ShapeNet objects with sparse point maps, summing the
field over the visible landmarks beats counting them by $+2.0$ points of
localization success rate (object-clustered 95\% interval $[+0.7, +3.2]$). It also beats
summing Harris-3D ranks by $+2.2$ $[+0.8, +3.5]$. On real indoor scenes from
SceneFun3D, our adaptation of the learned view scorer LWL falls below
counting on the test rooms. We therefore add the field to a learned
per-landmark planner. On the test rooms with sparse maps whose landmarks the
field also selects, the planner with the field beats the same planner
without it by $+1.8$ points. It beats the planner with Harris-3D or
curvature in its place by $+1.0$ and $+0.8$ points. All three
room-clustered intervals lie above zero. In both settings, the field helps
in sparse maps and is on par with Harris-3D in dense maps.
\end{abstract}

\section{Introduction}
\label{sec:intro}
\begin{figure*}[t]
  \centering
  \includegraphics[width=\textwidth]{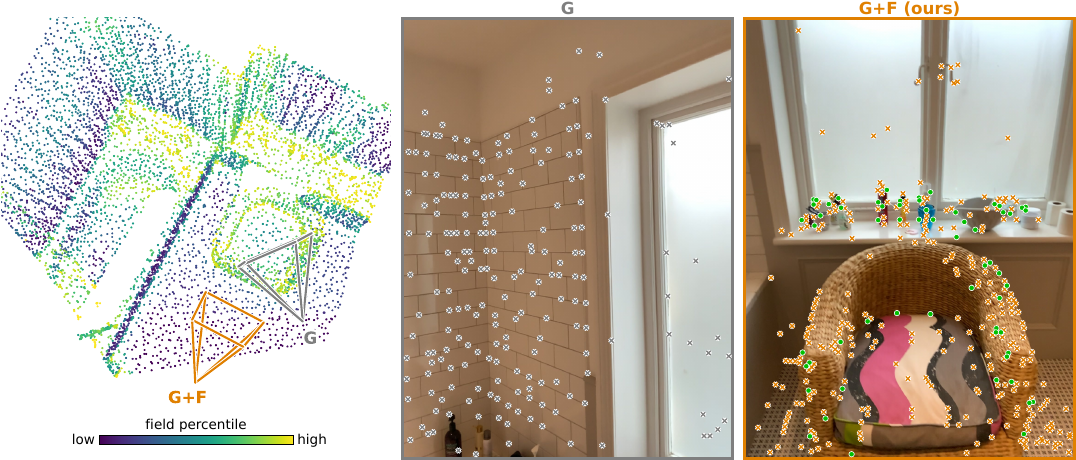}
  \caption{Problem setting, shown on one test decision in a real indoor
    scene. Before an image is taken, a planner ranks a few candidate camera
    poses using only the sparse map. The map points are coloured by their
    percentile of the image detectability field within the map. \arm{G}
    picks from geometric features alone, and \arm{G+F} also reads the field
    (Sec.~\ref{sec:perlm}). The two images were captured at the picked poses,
    and only the fixed 2D--3D localizer sees them. Green dots are matches to
    landmarks that PnP-RANSAC keeps as inliers, and crosses are the other
    matches. The \arm{G+F} pick localizes with 61 inliers ($2.2^\circ$,
    5\,cm). The \arm{G} pick yields 204 matches but no inlier, so no pose is
    found. The decision was selected by hand from the test decisions in which
    the \arm{G+F} pick localizes and the \arm{G} pick does not, to illustrate
    the difference.}
  \label{fig:teaser}
\end{figure*}

In 2D--3D localization, a camera estimates its pose by matching the
keypoints of an image to the points of a 3D map. The map can be a model of
a scene, which a camera uses to relocalize after it has lost track. The map
can also be the model of a known object, which a camera uses to estimate
the pose of that object. A camera held in a hand or mounted on a robot can
often turn or move a short distance before it takes the image. It can
therefore choose the viewpoint from which to take the image
(Figs.~\ref{fig:teaser}, \ref{fig:obj-setting}). Prior work learns to score such candidate viewpoints
on maps built by structure from
motion~\cite{hanlon2024active,digiammarino2024lwl,li2025actloc}.

The viewpoint is chosen before the image is taken, so the planner cannot
use the content of the image and has to rely on the map. Such a map usually
holds more than geometry. In a structure-from-motion map, each point also
carries a colour and the descriptors of the keypoints from which it was
triangulated. In this work, the planner reads only the geometry of the
map. This geometry is a point set and the depth needed to decide which
points a view sees. It contains no image, colour or descriptor. We then ask
how well geometry alone predicts whether localization succeeds. The
restriction applies only to the planner. Once the image is taken, the
localizer matches it against the descriptors of a reference image
(Secs.~\ref{sec:obj-protocol}, \ref{sec:localizer}). The same fixed
localizer serves every planner, so a difference in success comes only from
the chosen viewpoint.

With geometry alone, the planner has to predict where a 2D image detector
will find matchable keypoints without seeing the texture. In a
structure-from-motion map, the landmarks are triangulated detections, so
where they lie already shows where the texture is. Points sampled from a
model or a scan spread uniformly over the surface and carry no such signal
(Sec.~\ref{sec:exp-comp}). \fieldname{}~\cite{detfield} is an image
detectability field that provides this signal from geometry. From
the local geometry around a point, it predicts where the 2D image detector
SuperPoint~\cite{detone2018superpoint} fires on images of the surface. On
sparse maps, the landmarks that it selects localize better than those
selected by the classical 3D saliency score
Harris-3D~\cite{sipiran2011harris}, as reported in~\cite{detfield}. We use
the trained field as it is, without fine-tuning it for view selection. We
ask whether the field helps choose the viewpoint beyond counting landmarks
and other geometric cues. We also compare it with Harris-3D used in its
place.

We answer this question in two settings. The first setting uses synthetic
ShapeNet objects~\cite{chang2015shapenet} rendered as geometry-only images, and the
planner chooses among twelve viewpoints spread around a known object. Here
the planner uses the field directly as its utility, and the score of a
candidate is the sum of the field over its visible landmarks. The second
setting uses real indoor scenes from SceneFun3D~\cite{delitzas2024scenefun3d},
with a dense and a sparse map tier. In each decision, the candidates are
four to eight real frames from the same part of a room. Every candidate is
localized in advance, and the planners only rank the stored outcomes. We
retrain the learned view scorer LWL~\cite{digiammarino2024lwl} on the
outcomes of our localizer. On the test rooms, it falls below counting the
visible landmarks.
We therefore build a planner from geometry alone that beats counting. Like
the learned scorer of Hanlon et al.~\cite{hanlon2024active}, it describes
each visible landmark by geometric features, such as its distance and
viewing angle. A classifier predicts from these features whether the
landmark will be correctly matched, and the planner sums the predictions.
We add the field
to this planner and compare it with variants that add Harris-3D or the nine
curvature invariants that the field takes as input instead. In both
settings, we fixed the hypotheses and the analysis before running the test
set, and we ran the test set only once.

The two settings show the same pattern. On sparse objects, summing the
field picks better viewpoints than counting landmarks or summing Harris-3D
ranks. On dense objects, it does not differ significantly from either. In
sparse rooms with landmarks that the field selects among the map points,
adding the field raises the success rate of the planner. It raises the rate
more than Harris-3D or curvature added in its place. In dense rooms, the field,
Harris-3D and curvature are about equally useful. The value of the field
therefore lies in sparse maps. Our contributions are:
\begin{itemize}
  \item A view-selection protocol for 2D--3D localization in which the
    planner reads only the geometry of the map. The protocol covers pose
    estimation of known synthetic objects and relocalization in real indoor
    scenes, each with sparse and dense maps. In each setting, the test is
    registered before it is run and uses objects or rooms that neither the
    field nor the planner had seen. Under the indoor protocol, the published
    learned view scorer LWL~\cite{digiammarino2024lwl} falls below landmark
    counting on the test rooms (Secs.~\ref{sec:problem}, \ref{sec:exp-comp}).
  \item The image detectability field as a viewpoint utility. On objects,
    the planner uses the field directly. In rooms, the field is one input
    of the learned per-landmark planner. On sparse objects, summing the
    field raises the localization success rate by $+2.0$ points over counting and by $+2.2$
    points over summing Harris-3D ranks. On the test rooms with landmarks
    that the field selects among the sparse map points, adding the field
    raises the success rate of the planner by $+1.8$ points. The planner
    with the field also beats the planner with Harris-3D or curvature in
    its place by $+1.0$ and $+0.8$ points. All five clustered 95\%
    intervals lie above zero (Secs.~\ref{sec:exp-obj}, \ref{sec:exp-main}).
  \item On synthetic objects and in real indoor scenes alike, the value
    of the field lies in sparse maps. In both settings, the field beats
    Harris-3D on sparse maps. On dense maps, the field and Harris-3D do
    not differ significantly (Secs.~\ref{sec:exp-obj-dense},
    \ref{sec:exp-dense}).
\end{itemize}

Our claims rest on two registered tests. The first compares summing the
field with counting on sparse objects. The second compares the planner
variants on the test rooms in the sparse tier. A registered test on dense
objects serves as the dense counterpart of the first test. The other
results are descriptive. The design choices were made on the indoor
validation rooms, so the results on these rooms do not count as evidence
for our claims (Sec.~\ref{sec:experiments}).

\section{Related Work}
\label{sec:related}

\subsection{Viewpoint selection for visual localization}
Hanlon et
al.~\cite{hanlon2024active} select the viewpoint at a given location for
localization in a structure-from-motion landmark map. They compare
heuristics based on Fisher information, landmark count and viewing angle
with a learned scorer. LWL~\cite{digiammarino2024lwl} learns a viewpoint
scorer from the landmarks visible in a view, supervised by localizing
simulated images against the map. Among published learned view scorers,
LWL is the closest to our setting, because it reads only map geometry at
planning time. ActLoc~\cite{li2025actloc} encodes the landmarks of the map
and the mapping camera poses, and predicts localization accuracy over
viewing directions. Each landmark in ActLoc also carries its colour. Our
planner reads no colour, so we leave ActLoc out of the comparison.
Two recent works also model localization quality for
viewpoint planning. LIDIA~\cite{velentzas2026lidia} adds the direction of the
light source to a geometric model of localization quality and compares it
with the Fisher information field (FIF)~\cite{zhang2019fif,zhang2020fif}.
PIVOT~\cite{chen2026pivot} optimizes the viewing direction on the sphere to
maximize the visibility of features. These methods weight landmarks by
geometry, appearance or illumination. The image detectability field
weights landmarks by where a 2D image detector is predicted to fire.

\subsection{Perception-aware planning and localizability}
Davison and Murray~\cite{davison2002active} used active vision for
simultaneous localization and map building. Perception-aware planners score
future views by how much they help pose estimation. The scores come from
the photometric texture and the geometry of the
scene~\cite{costante2016perception}, from the Fisher information of the
landmarks~\cite{zhang2019fif,zhang2020fif} or from the uncertainty of the
scene and the camera~\cite{fang2022accurateacl}. Feature-selection methods
choose the visual cues or features that most improve pose
estimation~\cite{carlone2019attention,zhao2018goodfeature}. Map-quality
measures score a query pose from the structure of the landmark
map~\cite{merzic2017mapquality}. Most of these scores measure information
or uncertainty. The image detectability field instead predicts from
geometry alone where a 2D image detector fires.

\subsection{Detectability, matchability and mesh saliency}
Hartmann et al.~\cite{hartmann2014matchability} learn to predict from the
descriptor of a keypoint whether it will be matched.
MeshLoc~\cite{panek2022meshloc} localizes against dense meshes. Its results
remain competitive even when it renders the raw geometry without colour or
texture. In graphics, mesh saliency and best-view measures rate viewpoints
from curvature or from the visible
polygons~\cite{lee2005meshsaliency,feixas2009unified,bonaventura2018survey}.
These measures are designed for human viewing or shape understanding.
\fieldname{}~\cite{detfield} is instead trained to predict where
SuperPoint fires on images of a surface. For rooms it is trained on real
images, and for objects on rendered geometry-only images
(Sec.~\ref{sec:method}).
At query time, it reads only the local geometry and needs no descriptor or
image. In this work, we use the trained field as it is
for viewpoint selection.

\subsection{Active object recognition and pose estimation}
Next-best-view methods for object recognition and pose estimation choose the
next view to reduce the uncertainty of the class or the pose of the object.
They plan views for recognition from depth images~\cite{wu2015shapenets},
predict the uncertainty reduction of each view with a Hough
forest~\cite{doumanoglou2016recovering} or accumulate pose hypotheses over
RGB-D views~\cite{sock2017multiview}. These methods choose each view from
the observations made so far. Our object planner sees no image. It chooses
a single view before any observation, from the geometry of the object map
alone.

\begin{figure*}[t]
  \centering
  \includegraphics[width=\textwidth]{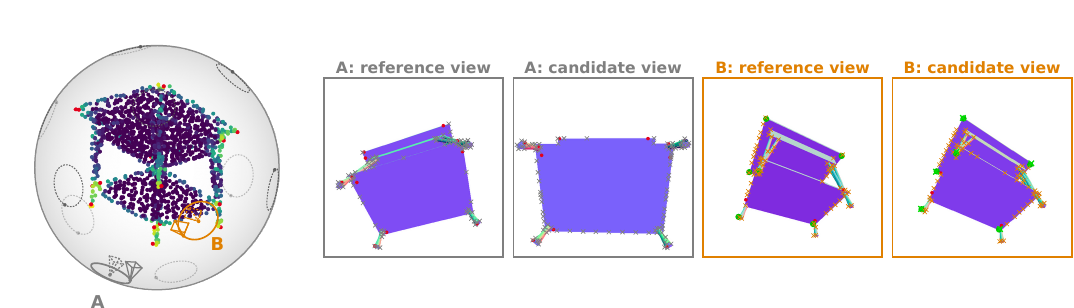}
  \caption{Experiment~1 on one table ($K = 16$ landmarks, voxel side
    $s = 0.04$, $\theta = 10^\circ$). Left: the sparse map inside a sphere
    of viewing directions, coloured by the rank of the field within the
    object. The landmarks are shown in red, and both planners use the same
    landmarks. Each solid frustum on a circle marks a candidate view, which
    lies $\theta$ from the reference view (dashed frustum) at the centre of
    the circle. A is the
    candidate that the planner counting visible landmarks picks, and B is the
    candidate that the planner summing the field picks, out of the twelve
    candidates of this decision. The other ten candidates are drawn as grey
    dots on dashed circles around their reference views. Right: the reference
    and candidate images of
    each side, cropped to the same window. Red dots are projections of the
    visible landmarks, and the candidate images use the true pose. Green dots
    are the landmarks that end up as PnP inliers, and bright green crosses are
    the candidate detections matched to them. The other crosses are
    SuperPoint detections. A sees mostly one flat face and gets no inlier. B
    sees the legs and the edges of the top and the shelf. It localizes with
    six inliers and is the only one of the twelve candidates that localizes.
    The images are shown as colour normal maps. The 2D detector sees the grey
    encoding of Sec.~\ref{sec:obj-protocol}. The example was selected by hand from the table decisions
    in which the field-sum pick localizes and the counting pick does not, to
    illustrate the difference.}
  \label{fig:obj-setting}
\end{figure*}

\section{Problem and Protocol}
\label{sec:problem}

\subsection{Setting}
\label{sec:setting}

The task is to choose a viewpoint for localization against a prior 3D map
(Fig.~\ref{fig:teaser}). Before the image is taken, a planner chooses among
a few candidate camera poses, and the image taken at the chosen pose gets
one localization attempt. We study this task in two experiments. The map is
a known synthetic object in Experiment~1 and a scanned real room in
Experiment~2 (Secs.~\ref{sec:obj-protocol}, \ref{sec:room-protocol}). The
planner reads only the geometry of the map. This
geometry is a point set $\mathcal{P}$ and the depth needed to decide which
points a view sees. The depth is rendered from the model of the object or
measured in the mapping frames $\mathcal{F}$ of the room. The planner reads
no image, colour or descriptor, including the candidate images. The
localizer is a fixed 2D--3D pipeline that matches the candidate image
against map points described by a reference image. We
restrict only the input of the planner, because we ask how much of the
localization success can be predicted from geometry alone.

A \emph{decision} $d$ offers a set of candidate poses $\mathcal{C}_d$ for
one localization attempt. The set has twelve or twenty candidates in the
object setting and four to eight in the indoor setting. A planner computes a utility $u(c)$ for
each candidate $c$ from $\mathcal{P}$, the pose of $c$ and the map points
that the planner finds visible from $c$. The planner
selects the set $\mathcal{A}_d = \argmax_{c \in \mathcal{C}_d} u(c)$ of
candidates with the highest utility. This set has more than one candidate
when utilities tie. Let $y(c) \in \{0,1\}$ indicate whether the image taken
at $c$ localizes successfully. The top-1 success of decision $d$ is
\begin{equation}
  s_d(u) = \frac{1}{|\mathcal{A}_d|} \sum_{c \in \mathcal{A}_d} y(c),
  \label{eq:top1}
\end{equation}
which is the expected success when ties are broken uniformly at random. We
report two references with every planner. The \meth{random} reference
$\frac{1}{|\mathcal{C}_d|}\sum_{c} y(c)$ is the success of a uniformly
random choice. The \meth{oracle} reference $\max_{c} y(c)$ needs the
outcomes and is an upper bound for every planner.

In both experiments, every candidate is localized in advance, and the
outcome is stored. The localization is repeated for each landmark budget,
map density and landmark selection (Secs.~\ref{sec:obj-protocol},
\ref{sec:room-protocol}). The planners only rank the stored candidates. All
planners are therefore scored on the same outcomes, which include the same
RANSAC draws, so every comparison can be paired by decision. This offline replay
selects views on rendered or recorded data, and no planner moves a camera.

\subsection{Experiment 1: synthetic objects}
\label{sec:obj-protocol}

\subsubsection{Task and data}
Experiment~1 studies model-based active pose estimation. The CAD model of
the object is known, and twelve reference views of the object are stored.
Each candidate view lies a fixed angle $\theta$ from one reference view,
which provides the descriptors for localizing it. The planner chooses which candidate to take
(Fig.~\ref{fig:obj-setting}). We use the ShapeNet~\cite{chang2015shapenet}
objects of the test set (Sec.~\ref{sec:registration}), which no earlier
experiment of this study used. Each object is scaled so that the longest
side of its bounding box is 0.9, and all lengths below are in these units.
The sparse map keeps one random point of the dense point cloud of the
object per voxel of side $s \in \{0.02, 0.04\}$, as in~\cite{detfield}.
We render images from the model at $640 \times 480$ pixels with a $60^\circ$
field of view. As in~\cite{detfield}, each image is a grey geometry-only
encoding of the surface normal and the depth.
The object is framed as in the renderings
on which \fieldname{} was trained.

\subsubsection{Decisions}
The twelve reference views look at the object from twelve directions, which a
Fibonacci lattice spreads evenly over a sphere. Around each reference
direction, eight candidate directions lie at an angle of exactly
$\theta \in \{10^\circ, 15^\circ\}$ from it. The eight directions are
evenly spaced on this circle, and each circle starts at its own random phase.
The phases are drawn with the index of the object as the seed. A decision
fixes the object, the voxel side $s$, the angle $\theta$ and an index $j$
among the eight directions. Its twelve candidates are the $j$-th directions
around the twelve reference directions. Every candidate lies exactly $\theta$
from its reference view, so the angle to the reference view cannot separate
the candidates. The candidates differ mainly in the side of the object that
they see. The 130 test objects thus give
$130 \times 2 \times 2 \times 8 = 4{,}160$ decisions.

\subsubsection{Localizer}
\label{sec:obj-localizer}
We use the object pipeline of~\cite{detfield} unchanged. The pipeline chooses
its landmarks among the sparse points by a score at each point. The two
landmark selections of our results differ only in this score. For the
\emph{field landmarks}, the score is the single-pass value of \fieldname{} at the
sparse point (Sec.~\ref{sec:field}). For the \emph{Harris landmarks}, the score is
Harris-3D~\cite{sipiran2011harris} on the sparse points, with 16 nearest
neighbours at $s = 0.02$ and 8 at $s = 0.04$. These values come
from~\cite{detfield}, which chose the best value for each voxel side on the
same objects. The sparse points are ranked by the score and thinned by 3D
non-maximum suppression with a radius of 0.05. The top
$K \in \{16, 32, 64\}$ points over the whole object become the landmarks, so
the landmark sets are nested in $K$.

For a pair of reference and candidate views, the landmarks visible in the reference view receive the
SuperPoint~\cite{detone2018superpoint} descriptors of the reference image at
their projections. A landmark is visible if its depth agrees with the
rendered depth within 0.03. SuperPoint detects at most 1024 keypoints in the
candidate image, with a 2D suppression radius of 4\,px and a score threshold
of 0.005. Each keypoint is matched to its nearest landmark descriptor with a
ratio test at 0.9. EPnP~\cite{lepetit2009epnp} inside
RANSAC~\cite{fischler1981random} then
estimates the pose with a reprojection threshold of 4\,px, confidence 0.999
and at most 2000 iterations. A pose needs at least four inliers. An attempt
succeeds if the rotation error is below $10^\circ$ and the camera-centre
error is below 0.05.

\subsubsection{Planner}
\label{sec:obj-planner}
The planner reads the sparse map, the landmarks and the values at the sparse
points that its utility needs, such as the field value or the Harris-3D rank.
The CAD model is known in this task, so the planner also reads depth rendered
from the model for the reference view and the candidate. It never sees the
images. With this depth, the planner applies the same visibility test as the
localizer, so its visibility is exact. In rooms, the planner has to estimate
visibility from the map (Sec.~\ref{sec:visibility}). The planner-visible
landmarks $\mathcal{V}(c)$ of a candidate $c$ are the top $K$ landmarks that
are visible in both the reference view and $c$. The scalar utilities
(Sec.~\ref{sec:scalar}) are computed from $\mathcal{V}(c)$, except
\meth{co-vis}, which ignores the landmarks.

\subsubsection{Statistics and registration}
\label{sec:obj-stats}
For each decision and planner, Eq.~\eqref{eq:top1} is computed for each $K$
and averaged over the three budgets. The success rate of a planner is the
mean of this score over the decisions. The decisions at both values of $s$
and $\theta$ are pooled. For a paired difference between two planners, we
resample objects with replacement 20{,}000 times (seed 0). In each resample,
we take the mean difference over the decisions of the drawn objects. The
95\% interval spans the middle 95\% of these means. We
cluster by object because the decisions of one object share its geometry. A
comparison passes if the lower end of its 95\% interval is above zero.
Intervals clustered by category are a secondary readout.

We committed the setting, the utilities, the analysis script and the
criterion before running the test, and ran it only once. The registered
criterion uses field landmarks, with $s$ and $\theta$ pooled. It requires
\meth{field-sum} to exceed \meth{count}, \meth{Harris-matched} and
\meth{permuted-field} (Sec.~\ref{sec:scalar}). The results with Harris
landmarks are reported with the same criterion. The comparison with
\meth{Harris-sum} and the results at single values of $s$ are reported
without a decision rule.

\subsubsection{Dense control}
A second registered test applies the same protocol to the dense point clouds
of the test objects. Each object has twenty reference directions, so each
decision offers twenty candidates. The budgets are $K \in \{32, 64, 128\}$,
and the 3D suppression radius is 0.02. Harris-3D uses its standard 32
nearest neighbours. The localizer and the success threshold are otherwise
the same. Its criterion is the same as for
sparse objects and uses field landmarks (Sec.~\ref{sec:exp-obj-dense}).

\subsection{Experiment 2: indoor relocalization}
\label{sec:room-protocol}

\subsubsection{Data and maps}
\label{sec:data}

Experiment~2 studies relocalization in real indoor scenes. The candidates of a
decision are real video frames near an anchor frame, and the planner chooses
which of them to take (Sec.~\ref{sec:decisions}).
For each room, SceneFun3D~\cite{delitzas2024scenefun3d} provides a laser
scan and handheld videos with camera poses, colour frames and depth maps. We
use one video per room and downscale its images by a factor of two. We
reconstruct a mesh from the laser scan with screened Poisson surface
reconstruction~\cite{kazhdan2013screened} at octree depth 11. The normals of
the scan points are oriented towards the cameras that see them, and the 3\%
of vertices with the lowest reconstruction density are removed. A \emph{crop} is the part of the mesh that a cluster of nearby
frames sees. The frames of a cluster have camera poses within $15^\circ$ and
0.5\,m of a seed frame.

Each crop gives two map tiers. The \emph{dense} map is a point cloud
sampled from the crop mesh. The \emph{sparse} map keeps one random point of
the dense map per voxel of side $\delta \in \{50, 100\}$\,mm, as in the sparse maps
of \fieldname~\cite{detfield}. The sparse tier is the registered main setting
(Sec.~\ref{sec:registration}). On sparse maps, the landmarks that
\fieldname{} selects localize better than those selected by
Harris-3D~\cite{detfield}. Voxel subsampling also makes the density nearly
uniform, so the density of the map carries no information on texture. The
planner then has to obtain this information from geometry.

The depth of the mapping frames is the high-resolution depth of the
dataset. We treat it as an ideal depth sensor. It only decides which map
points a mapping frame sees.

\subsubsection{Decisions and candidates}
\label{sec:decisions}

\begin{table}[t]
  \centering
  \caption{Experiment~2: protocol parameters}
  \label{tab:protocol}
  \footnotesize
  \begin{threeparttable}
  \begin{tabular}{@{}ll@{}}
    \toprule
    Quantity & Value \\
    \midrule
    Frame coverage (share of crop seen)    & $\ge 0.15$ \\
    Decisions per crop                     & $\le 30$ \\
    Separation of anchors                  & $\ge 0.25$\,m \\
    Candidate radius around the anchor     & $< 1.0$\,m \\
    Candidates per decision                & $4$ to $8$ \\
    Separation of candidate optical axes   & $\ge 12^\circ$ \\
    Time gap, mapping frame to candidates  & $> 0.5$\,s \\
    Reference window (rotation, distance)  & $< 20^\circ$, $< 0.6$\,m \\
    Landmark budget $K$                    & $64, 80, 96$ \\
    Sparse voxel size $\delta$             & $50, 100$\,mm \\
    PnP / RANSAC                           & EPnP, 8\,px, conf.\ 0.999 \\
    RANSAC iterations                      & $\le 2000$ \\
    Success                                & $< 5^\circ$ and $< 10$\,cm \\
    \bottomrule
  \end{tabular}
  \begin{tablenotes}[para,flushleft]\footnotesize
  \item[]\hspace{-\labelsep}The values are the same for both tiers and all splits.
  \end{tablenotes}
  \end{threeparttable}
\end{table}

We draw the decisions from all frames of the video of a crop
(Table~\ref{tab:protocol}). A frame is usable if it sees at least 15\% of a
fixed subsample of the dense points of the crop. Frames that see little of
the crop would only add trivial failures. Crops with fewer than 20 usable
frames are skipped. Decision generation uses the dense geometry in both
tiers. This geometry only sets up the decisions and is not an input of the
planner.

Two frames lie within the \emph{reference window} of each other if their
rotations differ by less than $20^\circ$ and their camera centres by less
than 0.6\,m. A frame is \emph{mapped} if another frame lies within its
reference window and was captured more than 0.5\,s apart. We visit the
usable frames in a seeded random order and take a frame as an anchor if it
is mapped and lies at least 0.25\,m from the earlier anchors. Each crop gives
at most 30 decisions. The candidates of a decision are mapped frames with
camera centres within 1.0\,m of the anchor. Starting from the anchor, we
pick up to eight of them greedily. Each new pick has the largest angle
between its optical axis and those already picked, and the picking stops
when this angle falls below $12^\circ$. The candidates thus differ in
orientation and by up to a metre in position. They are real frames with the
roll of the handheld camera. The mapping frames of a decision are the usable
frames captured more than 0.5\,s from every candidate. The video is dense,
so without this gap a mapping frame could duplicate a candidate. A candidate
is kept only if a mapping frame lies within its reference window, and a
decision only if four candidates and five mapping frames remain. The planner
thus chooses only among views that the map covers. A view of unmapped space
would fail for any planner and would not separate the planners.

Each candidate $c$ gets a \emph{reference frame} $r(c)$. Among the mapping
frames within the reference window of $c$, it is the one that shares the
most map points with $c$. The map points that $c$ sees come from the
visibility estimate of the planner (Sec.~\ref{sec:visibility}), and those
that a mapping frame sees come from its depth. This choice mimics retrieving
the closest mapped image from the map and the planned pose alone.

\subsubsection{Localizer}
\label{sec:localizer}

We use the room pipeline of~\cite{detfield} unchanged. It follows the object
pipeline of Experiment~1 (Sec.~\ref{sec:obj-localizer}) with three
differences. First, the landmarks are chosen in the reference frame. Among
the landmark positions visible in $r(c)$, the pipeline ranks the positions by
the landmark score, applies 3D non-maximum suppression and keeps the top $K$.
A position is visible if its depth agrees with the depth of $r(c)$ within
5\,cm. The suppression radius is 0.02 in the normalized coordinates of the
crop. We write $\mathcal{L}_K(r(c))$ for these \emph{landmarks}. Second, the
candidate image gives at most 4096 SuperPoint keypoints with a score above
0.002, and RANSAC uses a reprojection threshold of 8\,px. Third, an attempt
succeeds if the rotation and camera-centre errors with respect to the
dataset pose are below $5^\circ$ and 10\,cm. The other steps and parameters
are the same as for objects. Table~\ref{tab:protocol} lists the indoor
values.

The landmark budget is $K \in \{64, 80, 96\}$. Success depends on $K$, and
no single budget is canonical. We therefore average the results over the
three budgets within each decision (Sec.~\ref{sec:stats}).

The two landmark selections differ in the landmark score. For the main
\emph{field landmarks}, the score is the field, ranked as described in
Sec.~\ref{sec:field}. These landmarks form the sparse localization system of
\fieldname{} itself, which a view planner on a sparse map would work with.
At $\delta = 100$\,mm, the landmark positions include twelve off-sample
points per sparse point, since the field can be queried anywhere. These
points lie within $0.75\delta$ of the sparse point in its tangent plane and
within $\pm 0.25\delta$ along its normal. At 50\,mm, the landmark positions
are the sparse points. For the \emph{Harris landmarks}, the score is
Harris-3D~\cite{sipiran2011harris} on the sparse points with eight nearest
neighbours, so the field under test does not choose the landmarks. These
results are reported separately. The dense tier uses Harris landmarks only,
for the same reason.

\subsubsection{Planner-side visibility}
\label{sec:visibility}
The planner has no depth image of the candidate. It estimates from the map
alone which of the landmarks $\mathcal{L}_K(r(c))$ the candidate $c$ sees. In
the dense tier, it renders a z-buffer of the dense points. In the sparse
tier, it splats each sparse point as a square of the projected size of its
voxel (supplementary material, Sec.~S3). The landmarks that the planner
finds visible form $\mathcal{V}(c) \subseteq \mathcal{L}_K(r(c))$. All
utilities (Sec.~\ref{sec:method}) are computed from $\mathcal{V}(c)$, the
pose of $c$ and the map. They are computed for each $K$, $\delta$ and
landmark selection.

\subsubsection{Endpoint and statistics}
\label{sec:stats}

The endpoint is the binary success $y$ of one localization attempt. The
supplementary material (Sec.~S4) reports two other endpoints. For each
decision and planner, Eq.~\eqref{eq:top1} is computed for each pair of $K$
and $\delta$ and averaged over the six pairs. In the dense tier, it is
averaged over the three budgets. The paired differences and their 95\%
intervals are computed as on objects (Sec.~\ref{sec:obj-stats}), with rooms
in place of objects as clusters. We cluster by room because the decisions of
one room share its geometry.

\subsection{Splits and registration}
\label{sec:registration}

We split each dataset once and at random into a training set (70\%), a
validation set (10\%) and a test set (20\%). ShapeNet is split by object
within each category, over all objects of the 55 categories of
ShapeNetCore~\cite{chang2015shapenet}. SceneFun3D is split by room, over all
of its rooms. \fieldname{} and the learned planners use the same split, so
neither of them sees a test object or a test room before the final run. We
make every design choice of the planners on the validation set. Before the
run on the test set, we committed a registration. It fixes the commands, the
training data, the hyperparameters and seeds, the landmark selection, the
endpoint, the aggregation, the interval and the hypotheses
(Sec.~\ref{sec:hypotheses}). It also lists the quantities that are reported
without a decision rule. The test set is run only once, so the reported
results are not chosen among several runs.

\section{Method}
\label{sec:method}

\begin{figure*}[t]
  \centering
  \resizebox{\textwidth}{!}{%
  \begin{tikzpicture}[
      font=\footnotesize,
      box/.style={draw, rounded corners=2pt, align=center, inner sep=3pt,
                  minimum height=2.6em, fill=white},
      frz/.style={box, fill=black!8},
      trn/.style={box, very thick},
      fix/.style={box, dashed},
      arr/.style={-{Stealth[length=4pt]}, thin},
      lab/.style={font=\scriptsize, text=black!70}]
    \node[box, text width=1.6cm] (in)   at (0, 0)      {Map $\mathcal{P}$,\\ pose $c$};
    \node[box, text width=2.4cm] (phig) at (3.7, 0.75) {Geometric features\\ $\boldsymbol{\phi}_G(\ell, c)$};
    \node[frz, text width=2.4cm] (field) at (3.7,-0.75) {\fieldname};
    \node[trn, text width=2.3cm] (hgb)  at (8.85, 0)    {Classifier\\ $\hat p(z_\ell = 1 \mid \boldsymbol{\phi})$};
    \node[box, text width=2.3cm] (sum)  at (11.95, 0)   {$u(c) = \sum_{\ell \in \mathcal{V}(c)} \hat p$};
    \node[box, text width=1.9cm] (top)  at (14.85, 0)   {$\arg\max_c u(c)$};
    \node[fix, text width=1.6cm] (loc)  at (17.45, 0)   {Localizer};
    \draw[thin] (in.east) -- (1.35, 0);
    \draw[arr] (1.35, 0) |- (phig.west);
    \draw[arr] (1.35, 0) |- (field.west);
    \draw[thin] (phig.east) -- (6.75, 0.75) -- (6.75, 0);
    \draw[thin] (field.east) -- (6.75,-0.75) -- (6.75, 0);
    \node[lab, above] at (5.85,-0.75) {rank of $f(\ell)$};
    \draw[arr] (6.75, 0) -- (hgb.west);
    \draw[arr] (hgb) -- (sum);
    \draw[arr] (sum) -- (top);
    \draw[arr] (top) -- (loc);
  \end{tikzpicture}}
  \caption{Per-landmark learned planner of Experiment~2, drawn for the arm
    \arm{G+F}. The classifier is the only trained part, and it is trained on
    the training rooms. \fieldname{} is used as trained in~\cite{detfield},
    and the localizer is fixed. The other arms
    replace the field input by Harris-3D or curvature, or drop it
    (Sec.~\ref{sec:perlm}).}
  \label{fig:pipeline}
\end{figure*}
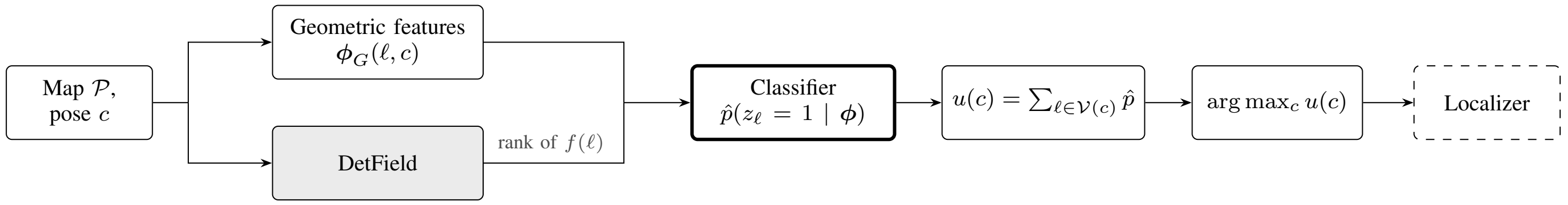

\subsection{The image detectability field}
\label{sec:field}

\fieldname~\cite{detfield} predicts the image detectability
$f(\mathbf{x} \mid \mathcal{P}) \in [0,1]$ at any query point $\mathbf{x}$
near the surface that the map $\mathcal{P}$ samples. Image detectability is
the response that a 2D image detector is expected to produce at the surface
point nearest to $\mathbf{x}$, aggregated over the views that see this
point. The input of \fieldname{} is geometry only. Each map point carries
nine curvature invariants. They are the squared Frobenius norm and the
squared determinant of the Hessian of a local quadric fit, and the surface
variation, each at 16, 32 and 64 neighbours and rank-normalized within the
map. An encoder reads the map. A decoder evaluates $f$ at the query point
from the encoder features and the invariants of the nearest map point.
\fieldname{} was trained in~\cite{detfield} to predict where SuperPoint fires.
For rooms, SuperPoint runs on real photographs of the scanned scene. For
objects, it runs on rendered grey images that encode the surface normal and
the depth. Colour enters only through the training target on real
photographs. It is not an input of the field. In the sparse rooms, we use the
weights that~\cite{detfield} trained for sparse maps, and in the dense rooms
its weights for dense maps. On objects, we use its object weights for both
the sparse and the dense maps. We take all weights from~\cite{detfield}
without further training. On objects, we query $f$ at the map points, as in
the object pipeline of~\cite{detfield}.

We read the field in two ways. Every planner below uses the single-pass
value $f(\mathbf{x})$. To rank the field landmarks in rooms
(Sec.~\ref{sec:localizer}), we follow~\cite{detfield} and average the
within-crop ranks of $f$ over four copies of the map. The copies are the
original map and three fixed random rotations of it, so that the ranking
depends less on the orientation of the map. On objects, the landmarks are
ranked by the single-pass value, as in the object pipeline
of~\cite{detfield}.

\subsection{Scalar planners}
\label{sec:scalar}

Each scalar planner reduces the planner-visible landmarks $\mathcal{V}(c)$
(Secs.~\ref{sec:obj-planner}, \ref{sec:visibility}) or the map to one
number. In Experiment~1, the scalar planners are the only planners, and the
field serves directly as the utility. In Experiment~2, they are baselines,
and the field is also an input of a learned planner (Sec.~\ref{sec:perlm}).
The last two planners below are placebos for objects. They keep the values
of the field and change where the values sit.
\begin{itemize}
  \item \meth{count}: $|\mathcal{V}(c)|$. This is the \emph{max} heuristic of
    Hanlon et al.~\cite{hanlon2024active}.
  \item \meth{co-vis} (objects and sparse rooms): the number of map points
    visible both from $c$ and from its reference view. It ignores the
    landmarks and measures the overlap with the reference image.
  \item \meth{field-sum}: $\sum_{\ell \in \mathcal{V}(c)} f(\ell)$. This
    planner sums the values $f(\ell) \in [0,1]$ of the field. It is the most
    direct use of the field as a view score.
  \item \meth{Harris-sum}: $\sum_{\ell \in \mathcal{V}(c)} \rho_H(\ell)$,
    where $\rho_H \in [0,1]$ is the rank of the Harris-3D response. The
    Harris-3D response has no common scale across maps, so this planner sums
    the ranks of the response. In rooms, the rank is taken among the landmark
    positions of the crop. On objects, it is taken among the sparse points of
    the object.
  \item \meth{angle}~\cite{hanlon2024active} (rooms): the number of
    landmarks in $\mathcal{V}(c)$ that were seen during mapping from a
    similar viewpoint. The original gives no thresholds, so we choose one of
    two readings on the validation set (supplementary material, Sec.~S3).
  \item \meth{permuted-field} (objects): \meth{field-sum} after the field
    values are randomly permuted among the sparse points of the object. It
    keeps the values and loses their placement, so it tests whether the
    utility depends on this placement.
  \item \meth{Harris-matched} (objects): \meth{field-sum} after each sparse
    point receives the field value of the same rank as its Harris-3D
    response. The values then follow the distribution of the field and the
    order of Harris-3D. This placebo was registered to separate the order of
    the values from the shape of their distribution.
\end{itemize}

\subsection{Per-landmark learned planner}
\label{sec:perlm}

We use this planner only in rooms (Experiment~2). Counting treats all
visible landmarks alike, but whether a landmark is matched depends on how
the candidate sees it. Following the per-landmark features of Hanlon et
al.~\cite{hanlon2024active}, we learn the probability that a visible landmark
yields a correct match. A candidate is scored by the sum of these
probabilities,
\begin{equation}
  u(c) = \sum_{\ell \in \mathcal{V}(c)} \hat p\big(z_\ell = 1 \mid
    \boldsymbol{\phi}(\ell, c)\big),
  \label{eq:perlm}
\end{equation}
which is the expected number of correct matches. A candidate with no visible landmark scores zero. The
label $z_\ell$ is 1 if a keypoint of the candidate image was matched to
$\ell$ and lies within 8\,px of the projection of $\ell$ under the true
pose. The localizer uses the same threshold for inliers.

The geometric features $\boldsymbol{\phi}_G$ use only the map, the reference
frame and the planned pose. For each landmark, they describe how $c$ sees
the landmark, how the view from $c$ relates to the view from $r(c)$, the
angles between the surface normal and the two viewing rays, and how many
landmarks and map points lie around it. The supplementary material
(Sec.~S3) lists all 13 features. Hanlon et al.\ describe each landmark by its
viewing range over all mapping images. Our features relate each landmark to
the reference image only, because the localizer matches only against that
image.

\emph{Arms.} Five planners share $\boldsymbol{\phi}_G$ and the classifier
(Fig.~\ref{fig:pipeline}). They differ only in the input added to
$\boldsymbol{\phi}_G$. \arm{G} adds nothing. \arm{G+F} adds the within-crop
rank of $f(\ell)$. \arm{G+H} adds $\rho_H(\ell)$. \arm{G+C9} adds the nine
invariants of the nearest map point. \arm{G+F+C9} adds the field rank and
the invariants together. \arm{G+H} tests whether Harris-3D adds as much as
the field. \arm{G+C9} tests whether the field adds anything beyond its own
input, since a classifier could learn from the invariants directly.

\emph{Training.} The classifier is a histogram gradient-boosted tree
ensemble, and the supplementary material (Sec.~S3) gives its
hyperparameters. We train one model per arm and landmark selection on the
training rooms. The models use only the records with $K = 96$, because the
landmark sets are nested in $K$ and smaller budgets would repeat the same
landmarks. At most $4 \times 10^5$ records are sampled (seed 0). The models
are applied unchanged to all $K$ and $\delta$, and the same models serve the
validation run and the test run. The dense tier uses the same recipe.

\emph{Why this base.} The registered plan was to add the field to
LWL~\cite{digiammarino2024lwl}, which is the closest published learned view
scorer (Sec.~\ref{sec:related}). Our adaptation of LWL did not exceed
\meth{count} on the validation set in either tier. A rule written before that
run then moved the hypotheses to the per-landmark planner. This planner was
the strongest geometry-only planner on the validation set.

\subsection{Published baselines}
\label{sec:competitors}

We adapt two published planners to our protocol.
LWL~\cite{digiammarino2024lwl} bins the landmarks on an image grid and feeds
per-bin statistics of their coordinates to an MLP. We retrain it on the
outcomes of our localizer. FIF~\cite{zhang2019fif,zhang2020fif} scores a view
by the Fisher information of the landmark bearings. For both, we choose the
variant on the validation set, and the supplementary material (Sec.~S3)
gives the details. The heuristics of Hanlon et al.~\cite{hanlon2024active}
are among the scalar planners (\meth{count}, \meth{angle}).

\subsection{Hypotheses}
\label{sec:hypotheses}

The registered criterion of Experiment~1 is given in
Sec.~\ref{sec:obj-stats}. The registered hypotheses of Experiment~2 compare
\arm{G+F} with \arm{G} and with the two arms in which another input replaces
the field input. They use field landmarks and the endpoint $y$, with $K$ and
$\delta$ averaged within each decision (Sec.~\ref{sec:stats}).
\begin{itemize}
  \item \textbf{H1}: \arm{G+F} $>$ \arm{G}. The field adds information to
    geometry.
  \item \textbf{H2}: \arm{G+F} $>$ \arm{G+H}. The field adds more than
    Harris-3D does.
  \item \textbf{H3}: \arm{G+F} $>$ \arm{G+C9}. The field adds more than the
    nine invariants that form its input.
\end{itemize}
Each hypothesis holds if the lower end of its room-clustered 95\% interval
is above zero. The main result passes only if all three hypotheses hold. A
secondary hypothesis repeats H1--H3 at $\delta = 100$\,mm and cannot change
the main decision. The results with Harris landmarks, at a single voxel size
and of the scalar planners are reported without a decision rule. The dense
tier has its own registration, which tests the same three hypotheses with
Harris landmarks. We report it as a second setting, and it cannot change the
main decision either.

\section{Experiments}
\label{sec:experiments}

\providecommand{\ci}[3]{$#1$\,[$#2$,\,$#3$]}
\providecommand{\cis}[3]{\ci{#1}{#2}{#3}$^{*}$}
\providecommand{\cisb}[3]{$\mathbf{#1}$\,[$#2$,\,$#3$]$^{*}$}

Our claims rest on two tests, each registered before a single run
(Secs.~\ref{sec:exp1}, \ref{sec:exp2}). All other results are descriptive.
We made the design choices on the indoor validation rooms, so the results on
these rooms do not count as evidence for our claims. The supplementary
material gives further analyses.

\subsection{Experiment 1: synthetic objects}
\label{sec:exp1}

Experiment~1 uses the field directly as the utility
(Sec.~\ref{sec:obj-protocol}). The registered test on sparse objects
carries the first claim, and the registered test on dense objects is its
dense counterpart.

\subsubsection{Sparse objects}
\label{sec:exp-obj}

\begin{table}[t]
  \centering
  \caption{Experiment~1: top-1 success (\%) of the scalar planners on objects}
  \label{tab:obj}
  \footnotesize
  \setlength{\tabcolsep}{4pt}
  \begin{threeparttable}
  \begin{tabular}{@{}lcccc@{}}
    \toprule
    & \multicolumn{2}{c}{Sparse objects} & \multicolumn{2}{c}{Dense objects} \\
    \cmidrule(lr){2-3}\cmidrule(l){4-5}
    Planner & Field lm. & Harris lm. & Field lm. & Harris lm. \\
    \midrule
    \meth{random}          & 16.0 & 3.7  & 30.6 & 18.4 \\
    \meth{co-vis}          & 22.9 & 5.3  & 36.7 & 25.0 \\
    \meth{count}           & 32.0 & 7.6  & 48.1 & 29.9 \\
    \meth{permuted-field}  & 27.9 & 6.8  & 44.3 & 29.6 \\
    \meth{Harris-matched}  & 29.6 & 7.9  & 44.0 & 30.2 \\
    \meth{Harris-sum}      & 31.8 & 7.8  & 47.8 & 30.3 \\
    \meth{field-sum}       & \textbf{34.0} & \textbf{9.0}  & \textbf{48.4} & \textbf{33.5} \\
    \midrule
    \meth{oracle}          & 75.3 & 27.1 & 93.1 & 74.8 \\
    \bottomrule
  \end{tabular}
  \begin{tablenotes}[para,flushleft]\footnotesize
  \item[]\hspace{-\labelsep}$K$ is averaged within each decision. Sparse: 130 test objects, with $s \in \{0.02, 0.04\}$ and $\theta \in \{10^\circ, 15^\circ\}$ pooled, $K \in \{16, 32, 64\}$ and 4{,}160 decisions per column. Dense control (Sec.~\ref{sec:exp-obj-dense}): 66 objects, with both values of $\theta$ pooled, $K \in \{32, 64, 128\}$ and 1{,}056 decisions per column. Field lm.\ and Harris lm.: landmarks chosen by \fieldname{} and by Harris-3D. A decision offers twelve candidates, and twenty in the dense control.
  \end{tablenotes}
  \end{threeparttable}
\end{table}

\begin{table*}[t]
  \centering
  \caption{Experiment~1: paired differences in top-1 success (percentage points) on objects}
  \label{tab:obj-effects}
  \footnotesize
  \setlength{\tabcolsep}{3pt}
  \begin{threeparttable}
  \begin{tabular}{@{}lccccc@{}}
    \toprule
    & \multicolumn{4}{c}{\meth{field-sum} minus} & \\
    \cmidrule(lr){2-5}
    Setting & \meth{count} & \meth{Harris-sum} & \meth{permuted-field} & \meth{Harris-matched}
      & \begin{tabular}[b]{@{}c@{}}\meth{Harris-matched}\\$-$\,\meth{Harris-sum}\end{tabular} \\
    \midrule
    \multicolumn{6}{@{}l}{\emph{Sparse objects (130 objects, 4{,}160 decisions)}} \\
    (i) Field lm., $s$ pooled (registered)  & \cisb{+2.0}{+0.7}{+3.2} & \cisb{+2.2}{+0.8}{+3.5} & \cisb{+6.0}{+4.3}{+7.7} & \cisb{+4.3}{+2.9}{+5.8} & \cis{-2.2}{-3.5}{-0.8} \\
    (ii) Field lm., $s = 0.02$              & \cisb{+2.7}{+0.9}{+4.5} & \cisb{+2.3}{+0.6}{+4.0} & \cisb{+7.2}{+4.6}{+9.7} & \cisb{+5.0}{+2.8}{+7.2} & \cis{-2.7}{-4.9}{-0.5} \\
    (iii) Field lm., $s = 0.04$             & \ci{+1.3}{-0.0}{+2.7} & \cisb{+2.0}{+0.4}{+3.7} & \cisb{+4.8}{+2.7}{+7.0} & \cisb{+3.6}{+1.8}{+5.4} & \cis{-1.6}{-2.9}{-0.3} \\
    (iv) Harris lm., $s$ pooled             & \cisb{+1.3}{+0.5}{+2.3} & \cisb{+1.2}{+0.3}{+2.2} & \cisb{+2.2}{+1.2}{+3.1} & \cisb{+1.1}{+0.2}{+2.1} & \ci{+0.1}{-0.3}{+0.5} \\
    \midrule
    \multicolumn{6}{@{}l}{\emph{Dense objects, control (66 objects, 1{,}056 decisions)}} \\
    (v) Field lm. (registered)              & \ci{+0.3}{-1.0}{+1.6} & \ci{+0.5}{-1.1}{+2.2} & \cisb{+4.1}{+1.3}{+7.0} & \cisb{+4.4}{+1.9}{+7.0} & \cis{-3.9}{-6.5}{-1.4} \\
    (vi) Harris lm.                         & \cisb{+3.6}{+0.5}{+6.9} & \cisb{+3.2}{+0.0}{+6.6} & \cisb{+3.9}{+1.0}{+6.8} & \cisb{+3.3}{+0.1}{+6.6} & \ci{-0.1}{-0.7}{+0.6} \\
    \bottomrule
  \end{tabular}
  \begin{tablenotes}[para,flushleft]\footnotesize
  \item[]\hspace{-\labelsep}The 95\% intervals come from an object-clustered bootstrap with 20{,}000 resamples. Rows (i) and (v) are the registered primary criteria, which require the first, third and fourth columns to exceed 0. Rows (iv) and (vi) are reported with the same criterion. All other entries are descriptive. lm.: landmarks.
  \item[$*$] The interval excludes 0. A lower end printed as $-0.0$ or $+0.0$ lies on the printed side of 0 before rounding.
  \end{tablenotes}
  \end{threeparttable}
\end{table*}

The registered criterion holds (Table~\ref{tab:obj-effects}, row~(i)).
Summing the field over the visible
landmarks raises the top-1 success from 32.0\% with \meth{count} to 34.0\%
(Table~\ref{tab:obj}). \meth{field-sum} exceeds \meth{count} by
\cis{+2.0}{+0.7}{+3.2}, \meth{permuted-field} by \cis{+6.0}{+4.3}{+7.7} and
\meth{Harris-matched} by \cis{+4.3}{+2.9}{+5.8}. With intervals clustered
by the 15 categories, the four differences of \meth{field-sum} in row~(i)
still exclude zero.

The \meth{Harris-matched} placebo is distorted in row~(i). It falls below
plain \meth{Harris-sum} (\cis{-2.2}{-3.5}{-0.8}). The long-tailed values of
the field are handed out in Harris-3D order, so a few points that Harris-3D
ranks high and the field ranks low dominate the sum. The margin of $+4.3$
therefore overstates the advantage of the field order. We read the
comparison with \meth{Harris-sum} as the contrast with Harris-3D instead,
and \meth{field-sum} exceeds \meth{Harris-sum} by \cis{+2.2}{+0.8}{+3.5}.
The permuted field
falls below \meth{count} (27.9\% against 32.0\%), so the gain depends on
the placement of the field values. Table~\ref{tab:obj-effects} also lists the results
with Harris landmarks and at each voxel size. The supplementary material
(Sec.~S1) discusses them, together with a second grouping of the decisions.

\subsubsection{Dense objects}
\label{sec:exp-obj-dense}

With field landmarks, the registered criterion fails
(Table~\ref{tab:obj-effects}, row~(v)). The differences of \meth{field-sum}
to \meth{count} and to \meth{Harris-sum} are \ci{+0.3}{-1.0}{+1.6} and
\ci{+0.5}{-1.1}{+2.2}, so it is level with both. It still beats the two
placebos. As on sparse
objects, \meth{Harris-matched} is distorted and falls below
\meth{Harris-sum}. With field landmarks, the field thus beats counting
and the Harris-3D order on sparse objects, and it is level with them on
dense objects. The supplementary material (Sec.~S2) describes
how this test came about.

\subsection{Experiment 2: indoor relocalization}
\label{sec:exp2}

Experiment~2 adds the field to the per-landmark planner
(Sec.~\ref{sec:room-protocol}). The registered comparison on the test rooms
in the sparse tier carries the second claim, and the dense tier is a second
setting.

\subsubsection{Data}
\label{sec:exp-data}

Of the 54 test rooms, 49 produced crops, and the other five produced none.
The 49 rooms gave 378 crops, and 367 of them yielded decisions. Ten crops
had no usable anchor, and one had fewer than 20 usable frames
(Sec.~\ref{sec:decisions}). In the sparse tier, the test set has 4{,}385
decisions and 154{,}158 localization records with field landmarks, one for
each candidate, budget and voxel size. In the dense tier, it has 4{,}412
decisions in the same 49 rooms. The two tiers draw decisions with the same
code but with different seeds, so their decisions are not identical. The
validation results below come from 12 rooms, with 1{,}912 decisions in the
sparse tier and 1{,}913 in the dense tier.

\subsubsection{Sparse maps}
\label{sec:exp-main}

\begin{table}[t]
  \centering
  \caption{Experiment~2: top-1 success (\%) of all planners in the sparse tier}
  \label{tab:planners}
  \footnotesize
  \setlength{\tabcolsep}{4pt}
  \begin{threeparttable}
  \begin{tabular}{@{}lccccc@{}}
    \toprule
    & \multicolumn{2}{c}{Test, field lm.} & \multicolumn{2}{c}{Val., field lm.} & Test, Harris lm. \\
    \cmidrule(lr){2-3}\cmidrule(lr){4-5}\cmidrule(l){6-6}
    Planner & top-1 & Rel. & top-1 & Rel. & top-1 \\
    \midrule
    \meth{random}     & 31.3 & 40.2 & 35.9 & 44.6 & 28.2 \\
    \meth{count}      & 45.3 & 58.2 & 47.8 & 59.5 & 39.0 \\
    \meth{co-vis}     & 44.8 & 57.6 & 51.0 & 63.4 & 40.4 \\
    \meth{angle}      & 46.8 & 60.1 & 49.8 & 61.9 & 38.2 \\
    \meth{Harris-sum} & 44.4 & 57.0 & 47.9 & 59.6 & 42.3 \\
    \meth{field-sum}  & \textbf{49.0} & \textbf{63.0} & \textbf{51.4} & \textbf{64.0} & \textbf{43.6} \\
    \midrule
    \arm{G}           & 50.2 & 64.5 & 54.8 & 68.2 & 44.6 \\
    \arm{G+H}         & 51.1 & 65.7 & 55.4 & 68.9 & 46.5 \\
    \arm{G+C9}        & 51.3 & 65.8 & 55.6 & 69.2 & 46.1 \\
    \arm{G+F}         & \textbf{52.1} & \textbf{66.9} & \textbf{56.1} & \textbf{69.8} & \textbf{46.6} \\
    \arm{G+F+C9}      & \textbf{52.3} & \textbf{67.2} & \textbf{56.5} & \textbf{70.3} & \textbf{46.7} \\
    \midrule
    \meth{oracle}     & 77.9 & --   & 80.4 & --   & 74.0 \\
    \bottomrule
  \end{tabular}
  \begin{tablenotes}[para,flushleft]\footnotesize
  \item[]\hspace{-\labelsep}Endpoint $y$, with $K$ and $\delta$ averaged within each decision. Rel.: top-1 success divided by the success of the oracle (\%), which is 100 for the oracle by definition. Field lm.\ and Harris lm.: landmarks chosen by \fieldname{} and by Harris-3D. Field landmarks are the registered setting, and Harris landmarks are reported without a decision rule. Test: 49 rooms, 4{,}385 decisions. Validation: 12 rooms, 1{,}912 decisions. LWL and FIF are in Table~\ref{tab:competitors}.
  \end{tablenotes}
  \end{threeparttable}
\end{table}

\begin{table*}[t]
  \centering
  \caption{Experiment~2: paired differences in top-1 success (percentage points)}
  \label{tab:effects}
  \footnotesize
  \setlength{\tabcolsep}{3pt}
  \begin{threeparttable}
  \begin{tabular}{@{}lccccc@{}}
    \toprule
    Setting & H1: \arm{G+F}$-$\arm{G} & H2: \arm{G+F}$-$\arm{G+H} & H3: \arm{G+F}$-$\arm{G+C9}
      & \arm{G+F+C9}$-$\arm{G+C9} & \arm{G}$-$\meth{count} \\
    \midrule
    \multicolumn{6}{@{}l}{\emph{Test set, sparse tier}} \\
    (a) Field lm., merged (main)        & \cisb{+1.8}{+1.3}{+2.4} & \cisb{+1.0}{+0.5}{+1.5} & \cisb{+0.8}{+0.3}{+1.4} & \cisb{+1.0}{+0.6}{+1.5} & \cis{+5.0}{+2.8}{+7.2} \\
    (b) Field lm., 100\,mm (secondary)  & \cisb{+1.1}{+0.5}{+1.8} & \ci{+0.5}{-0.1}{+1.1} & \ci{+0.4}{-0.1}{+1.1} & \cisb{+0.7}{+0.2}{+1.3} & \cis{+4.1}{+2.1}{+6.0} \\
    (c) Field lm., 50\,mm               & \cisb{+2.6}{+1.9}{+3.3} & \cisb{+1.4}{+0.9}{+2.1} & \cisb{+1.2}{+0.6}{+1.9} & \cisb{+1.3}{+0.7}{+2.0} & \cis{+5.8}{+3.3}{+8.6} \\
    (d) Harris lm., merged              & \cisb{+1.9}{+1.1}{+2.9} & \ci{+0.1}{-0.5}{+0.7} & \cisb{+0.4}{+0.0}{+0.9} & \cisb{+0.6}{+0.2}{+1.0} & \cis{+5.6}{+3.8}{+7.4} \\
    \midrule
    \multicolumn{6}{@{}l}{\emph{Validation set}} \\
    (e) Sparse, field lm., merged       & \cisb{+1.3}{+0.1}{+2.8} & \ci{+0.7}{-0.1}{+1.8} & \ci{+0.5}{-0.5}{+1.6} & \ci{+0.9}{-0.0}{+1.9} & \cis{+7.0}{+3.0}{+10.6} \\
    (f) Dense                           & \cisb{+2.3}{+0.4}{+4.1} & \ci{+0.8}{-1.1}{+2.7} & \ci{+0.3}{-0.6}{+1.4} & \ci{+0.7}{-0.1}{+1.6} & \cis{+4.4}{+1.1}{+8.4} \\
    \midrule
    \multicolumn{6}{@{}l}{\emph{Dense tier, test set (descriptive)}} \\
    (g) Dense, test set               & \cisb{+3.5}{+2.6}{+4.4} & \ci{+0.3}{-0.4}{+1.1} & \ci{-0.6}{-1.2}{+0.0} & \cisb{+0.8}{+0.4}{+1.2} & \ci{+0.3}{-1.9}{+3.0} \\
    \bottomrule
  \end{tabular}
  \begin{tablenotes}[para,flushleft]\footnotesize
  \item[]\hspace{-\labelsep}The 95\% intervals come from a room-clustered bootstrap with 20{,}000 resamples. The point estimate is paired per decision and computed before rounding, so it can differ by 0.1 from the difference of the rounded rates in Table~\ref{tab:planners}. Row~(a) is the registered main decision, and row~(b) the registered secondary hypothesis. All other rows are descriptive. The sparse rows of the test set use the same 4{,}385 decisions. Rows (a)--(c) use field landmarks, and row~(d) uses Harris landmarks. The dense tier always uses Harris landmarks, and there \arm{G+F} reads the field of the dense map. lm.: landmarks.
  \item[$*$] The interval excludes 0. A lower end printed as $+0.0$ is positive before rounding.
  \end{tablenotes}
  \end{threeparttable}
\end{table*}

\begin{figure}[t]
  \centering
  \includegraphics[width=\columnwidth]{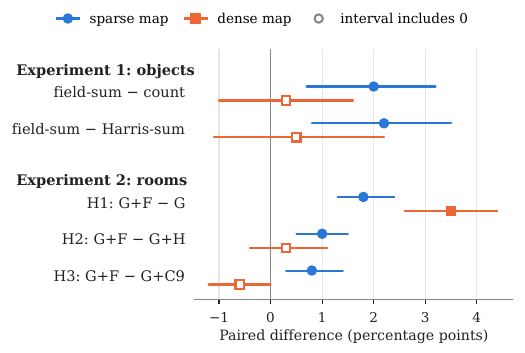}
  \caption{Paired differences in top-1 success with 95\% intervals, for
  sparse and dense maps. Experiment~1 uses field landmarks on objects, with
  intervals clustered by object. Experiment~2 uses the test rooms, with field
  landmarks in the sparse tier and Harris landmarks in the dense tier, and
  intervals clustered by room. The \meth{field-sum}$-$\meth{Harris-sum} rows
  are descriptive. All other rows are registered tests, and the test on
  dense objects fails.}
  \label{fig:forest}
\end{figure}

Table~\ref{tab:planners} lists the success rates of all planners, and
Table~\ref{tab:effects} gives the paired differences. Row~(a) of
Table~\ref{tab:effects} holds the registered comparisons
(Sec.~\ref{sec:hypotheses}). On the test set, adding the field to the
per-landmark planner raises the top-1 success from 50.2\% to 52.1\%.
\arm{G+F} exceeds \arm{G} by \cis{+1.8}{+1.3}{+2.4}, the Harris-3D arm
\arm{G+H} by \cis{+1.0}{+0.5}{+1.5} and the curvature arm \arm{G+C9} by
\cis{+0.8}{+0.3}{+1.4}. All three registered hypotheses hold. The
intervals on the test set are about half as wide as on the validation set
(row~(e)). The field also adds to the curvature arm when it is given on top of it
(\arm{G+F+C9}$-$\arm{G+C9}: \cis{+1.0}{+0.6}{+1.5}).

In row~(a), the largest difference is that of the learned geometric planner
over counting, \arm{G}$-$\meth{count} \cis{+5.0}{+2.8}{+7.2}. The field adds
to this gain. Fig.~\ref{fig:forest} shows the registered comparisons in all
settings.

The supplementary material (Sec.~S4) reports the other analyses of this
run. They cover the secondary hypothesis at 100\,mm, Harris landmarks, the
scalar planners and two other endpoints, and none of them changes the
registered decision.

\begin{figure*}[t]
  \centering
  \includegraphics[width=\textwidth]{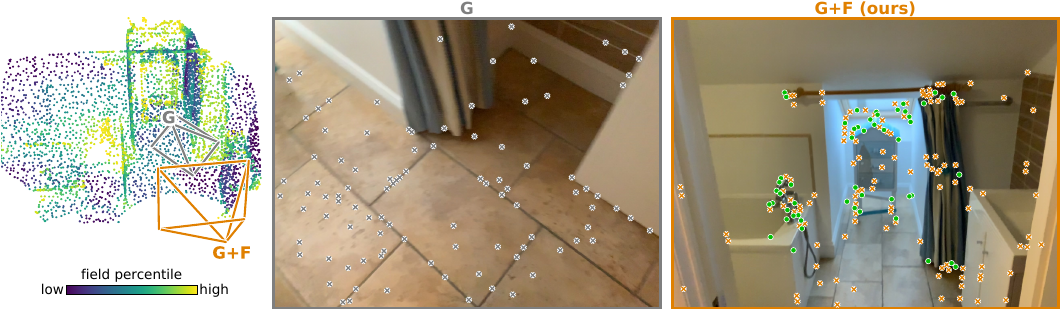}
  \caption{A second test decision in a real indoor scene, drawn as in
  Fig.~\ref{fig:teaser}. The \arm{G} pick faces the floor tiles and yields
  97 matches but no inlier, so no pose is found. The \arm{G+F} pick
  localizes with 52 inliers ($0.8^\circ$, 4\,cm). The decision was
  selected by hand from the same pool as Fig.~\ref{fig:teaser}, to
  illustrate the difference.}
  \label{fig:qual}
\end{figure*}

Fig.~\ref{fig:qual} shows a second test decision in which the \arm{G+F}
pick localizes and the \arm{G} pick does not.

\subsubsection{Dense maps}
\label{sec:exp-dense}

In the dense tier, every visible landmark has dense geometry around it. On
the test set, adding the field to \arm{G} raises the success by
\cis{+3.5}{+2.6}{+4.4}, so H1 holds (Table~\ref{tab:effects}, row~(g)).
Adding Harris-3D or the curvature invariants helps about as much
(\arm{G+H}$-$\arm{G} \cis{+3.2}{+2.2}{+4.1}, \arm{G+C9}$-$\arm{G}
\cis{+4.0}{+2.9}{+5.3}), so H2 and H3 do not hold. On dense maps, the field
thus adds about as much as Harris-3D and curvature. The validation set
shows the same pattern with wider intervals (row~(f)). The success rates of
all planners in this tier are in the supplementary material (Sec.~S5).

Together with Sec.~\ref{sec:exp-main}, these results place the specific
value of the field in sparse maps. On sparse maps, the field adds to
geometry and adds more than Harris-3D and curvature. On dense maps, the
three saliency inputs serve about equally well. The object experiments show
the same pattern (Sec.~\ref{sec:exp-obj-dense}).

\subsubsection{Published baselines}
\label{sec:exp-comp}

\begin{table}[t]
  \centering
  \caption{Experiment~2: published baselines, top-1 success (\%) and difference to \meth{count} (percentage points)}
  \label{tab:competitors}
  \footnotesize
  \setlength{\tabcolsep}{4pt}
  \begin{threeparttable}
  \begin{tabular}{@{}llccc@{}}
    \toprule
    & & Val. & \multicolumn{2}{c}{Test} \\
    \cmidrule(l){4-5}
    Tier & Planner & top-1 & top-1 & $-$\,\meth{count} \\
    \midrule
    Dense  & LWL, all points$^{\dagger}$ & 52.7 & 50.3 & \cis{-10.0}{-12.9}{-6.9} \\
    Dense  & LWL, localizer lm.          & 60.9 & 57.0 & \cis{-3.4}{-5.3}{-1.4} \\
    Dense  & FIF                         & 59.3 & 55.1 & \cis{-5.3}{-9.8}{-1.2} \\
    \midrule
    Sparse & LWL, all points$^{\dagger}$ & 39.9 & 37.0 & \cis{-8.3}{-10.6}{-6.1} \\
    Sparse & FIF                         & 39.6 & 38.3 & \cis{-7.0}{-9.9}{-3.9} \\
    \bottomrule
  \end{tabular}
  \begin{tablenotes}[para,flushleft]\footnotesize
  \item[]\hspace{-\labelsep}Endpoint $y$ (Sec.~\ref{sec:competitors}). The differences to \meth{count} are paired per test decision and come with room-clustered 95\% intervals. Test: 49 rooms, 4{,}412 dense and 4{,}385 sparse decisions. Val.: 12 validation rooms. Dense: Harris landmarks, \meth{count} 60.4\% on the test set and 64.8\% on the validation set. Sparse: field landmarks, \meth{count} 45.3\% and 47.8\%. The LWL family is either the landmarks that the localizer matches (lm.) or all planner-visible map points (all points).
  \item[$*$] The interval excludes 0.
  \item[$\dagger$] The registered family, chosen on the validation set.
  \end{tablenotes}
  \end{threeparttable}
\end{table}

On the test rooms, neither published baseline beats \meth{count} in either
tier (Table~\ref{tab:competitors}). The table reports our adaptation of
LWL~\cite{digiammarino2024lwl} to this protocol (supplementary material,
Sec.~S3), so its numbers are not comparable with the published numbers of
LWL. The supplementary material (Sec.~S3) describes how the maps of the two
protocols differ. FIF~\cite{zhang2019fif} falls below counting in
both tiers. Hanlon et al.~\cite{hanlon2024active} also rank their
Fisher-information heuristic below counting.

\section{Discussion and Limitations}
\label{sec:discussion}

\emph{Why the gain is small.} A diagnosis on the validation set
(supplementary material, Sec.~S6) suggests that image detectability is not
the main bottleneck of this task. The field carries some information on
whether a visible landmark is matched correctly. However, the utility of a
candidate sums over tens of landmarks, and the geometric features already
separate most candidates. Whether a correct match becomes a RANSAC inlier
also depends more on geometry than on the field. The other uses of the
field that we tried there did not enlarge the gain, so we use the field only
as a per-landmark input.

\emph{Scope of the claim.} We state the advantage over Harris-3D only for
landmarks that the field selects. With Harris landmarks, the results do not
follow the same pattern (supplementary material, Secs.~S1, S4). In sparse
rooms with Harris landmarks, the field arm does not beat the Harris-3D arm.
The advantage therefore belongs to the landmark selection of \fieldname{}
and the planner together. We have one explanation for the difference
between sparse and dense maps, which we have not tested. On dense geometry,
Harris-3D and
the curvature invariants are computed from many neighbours and already
carry most of the information that the field computes from the same
invariants. On sparse samples, these
per-point estimates are coarser, while the encoder of the field reads the
map around the point.

\emph{Limitations.} The object experiment is synthetic. Its images are
geometry-only images rendered from the model, and the object weights of the
field were trained on the same kind of image. Its planner also sees exact
visibility from the model. The experiment therefore shows what the field can
do in a clean setting, and Experiment~2 tests the field on real captures.
Indoors, the mapping frames see the map through the high-resolution depth of
the dataset, which acts as an ideal depth sensor. This depth only decides
which map points a mapping frame sees, and the planner sees no depth of the
candidate. All evidence on real images comes from one dataset, SceneFun3D.
On two scenes from HM3D~\cite{ramakrishnan2021hm3d}, a descriptive check also finds that adding the field to
\arm{G} raises the success (supplementary material, Sec.~S7).

Each decision offers at most eight candidates, because the candidates are
real frames. Prior protocols render far more
candidates~\cite{hanlon2024active,digiammarino2024lwl}. In return, real
frames carry no rendering gap. Every result is an offline replay of recorded
captures. The planner ranks stored outcomes and never moves a camera, so we
do not test closed-loop behaviour. The replay scores all planners on the
same outcomes. The localizer was fixed by the registration, and the results
hold for this localizer.

\section{Conclusion}
\label{sec:conclusion}

We asked whether an image detectability field is a useful viewpoint utility
for a planner that reads only the geometry of the map. We answered in two
settings, each with a test registered before the runs. On synthetic sparse
objects, summing the field over the visible landmarks raised the
localization success rate by $+2.0$ points over counting them and by $+2.2$
points over summing Harris-3D ranks. On dense objects, it was on par with
both. On the test rooms with sparse maps, the field also selects the
landmarks. Adding the field to a learned per-landmark planner raised the
success rate by $+1.8$ points over the same planner without it. It also beat
the planner with Harris-3D or curvature in its place by $+1.0$ and $+0.8$
points. This gain over Harris-3D depends on the landmarks that the field
selects. On dense maps, the field added about as much as Harris-3D and
curvature. In both settings, the value of the field lies in sparse maps.
The diagnosis in Sec.~\ref{sec:discussion} suggests that a larger gain needs
a view-dependent signal, for instance a field conditioned on the viewing
direction and the distance. Testing the object result on real images and
with a learned planner is another step for future work.

\bibliographystyle{IEEEtran}
\bibliography{refs}

\end{document}


\bstctlcite{IEEEexample:BSTcontrol}

\title{Supplementary Material:\\ An Image Detectability Field as a
Viewpoint Utility for 2D--3D Localization in Sparse Maps}
\author{Han Sun, Jun Li and Christy Jie Liang}
\maketitle

This document gives details and analyses that support the
main paper. Sections~S1 and S2 concern Experiment~1 on ShapeNet
objects~\cite{chang2015shapenet}, and Sections~S3 to S7 concern
Experiment~2 on real indoor scenes from SceneFun3D~\cite{delitzas2024scenefun3d}.
Field landmarks and Harris landmarks are the landmarks chosen by
\fieldname{}~\cite{detfield} and by Harris-3D~\cite{sipiran2011harris}.
Section and table numbers without the prefix S refer
to the main paper. Intervals are 95\% intervals from a room-clustered bootstrap
with 20{,}000 resamples unless stated otherwise; $^{*}$: the interval
excludes 0.

\section{Experiment 1: further analyses}
\label{sec:s-obj}

This section covers rows (ii)--(iv) and (vi) of Table~III and a second
grouping of the decisions.

\emph{Harris landmarks.} With Harris landmarks, the criterion of row~(i)
also holds on sparse objects (Table~III, row~(iv)). \meth{field-sum}
exceeds \meth{Harris-sum} by \cis{+1.2}{+0.3}{+2.2}. The
\meth{Harris-matched} placebo is not distorted here. It is level with
\meth{Harris-sum} (\ci{+0.1}{-0.3}{+0.5}). All rates with Harris landmarks
are low. \meth{field-sum} succeeds in 9.0\% of the decisions,
\meth{random} in 3.7\% and the \meth{oracle} in 27.1\% (Table~II). On
dense objects, the criterion of row~(v) also holds with Harris landmarks
(Table~III, row~(vi)), and \meth{field-sum} exceeds \meth{Harris-sum} by
\cis{+3.2}{+0.0}{+6.6}.

\emph{Voxel size.} With field landmarks, all four differences of
\meth{field-sum} exclude zero at the voxel side $s = 0.02$ (row~(ii)). At
$s = 0.04$, the margin over \meth{count} shrinks to \ci{+1.3}{-0.0}{+2.7}.
The margins over \meth{Harris-sum} and over the two placebos still exclude
zero (row~(iii)). With Harris landmarks at $s = 0.04$, \meth{field-sum}
succeeds in 3.7\% of the decisions, and its margin over \meth{count} is
\ci{+0.2}{-0.4}{+0.9}. Even the \meth{oracle} succeeds in only 15.1\% of
these decisions.

\emph{Decisions around one reference.} In a second grouping, the eight
directions around one reference direction form a decision
(Sec.~III-B2). This grouping gives 6{,}240 decisions. The candidates of
each decision see nearly the same side of the object. With field
landmarks, the six planners succeed in 17.4\% to 18.2\% of these decisions,
against 16.0\% for \meth{random} and 47.7\% for the \meth{oracle}.
\meth{field-sum} is level with \meth{count} (\ci{-0.1}{-0.3}{+0.2}). This
suggests that directions at the same angle around one reference are hard
to tell apart from geometry. On objects, the field helps to choose the
side of the object to look at. It does not help to choose among the
directions around one reference.

\section{Experiment 1: an earlier dense-object test}
\label{sec:s-ring}

A test on 60 dense objects came before the dense-object test of
Sec.~V-A2. Its registration grouped the decisions by mistake. The eight
candidate directions around one reference formed a decision. We had
intended the grouping of Sec.~III-B2, in which a decision takes one
candidate around each of the twelve reference directions. Under the
registered grouping, the candidates of a decision see nearly the same side
of the object. With both landmark selections, all planners except
\meth{random} were within one point of each other. With field landmarks,
\meth{field-sum} differed from \meth{count}, \meth{Harris-matched} and
\meth{permuted-field} by $-0.2$, $-0.2$ and $+0.5$ points. The test
therefore did not separate the planners and decided nothing about the
field.

We then registered the grouping of Sec.~III-B2 as a new test, which is the
dense-object test of Sec.~V-A2. Its 20 reference directions do not
coincide with the earlier twelve, and its candidate directions start at
new random phases. This test uses 66 objects, and 60 of them had been used
in the earlier test. It is therefore weaker evidence than the
sparse-object test of Sec.~V-A1. That test was the third registered object
test, and its 130 objects had not been used in any earlier test of view
selection.

\section{Experiment 2: protocol and planner details}
\label{sec:s-protocol}
\begin{table*}[t]
  \centering
  \caption{Experiment~2: our protocol and prior single-step view selection for localization}
  \label{tab:settings}
  \footnotesize
  \setlength{\tabcolsep}{4pt}
  \begin{threeparttable}
  \begin{tabular}{@{}>{\raggedright\arraybackslash}p{0.10\textwidth}
                    >{\raggedright\arraybackslash}p{0.21\textwidth}
                    >{\raggedright\arraybackslash}p{0.19\textwidth}
                    >{\raggedright\arraybackslash}p{0.14\textwidth}
                    >{\raggedright\arraybackslash}p{0.13\textwidth}
                    >{\raggedright\arraybackslash}p{0.14\textwidth}@{}}
    \toprule
    Work & Map read by the planner & Candidate views & Localizer & Success threshold & Data \\
    \midrule
    Hanlon et al.~\cite{hanlon2024active}
      & SfM landmarks with their mapping observations, a TSDF for occlusion
      & Rendered, 50 directions of yaw and pitch at each of 100 waypoints
      & hloc~\cite{sarlin2019hloc} with SuperPoint~\cite{detone2018superpoint} or SIFT~\cite{lowe2004sift}
      & Six, from 0.025\,m\,/\,$1^\circ$ to 1\,m\,/\,$5^\circ$
      & HM3D~\cite{ramakrishnan2021hm3d}, 9 scenes \\ \addlinespace[3pt]
    LWL~\cite{digiammarino2024lwl}
      & SfM points triangulated from SIFT features
      & Rendered, grid positions with sphere-sampled directions, no roll
      & COLMAP~\cite{schoenberger2016sfm} with SIFT
      & Four, from 0.05\,m\,/\,$0.4^\circ$ to 5\,m\,/\,$10^\circ$
      & HM3D, 10 training and 2 test scenes \\ \addlinespace[3pt]
    ActLoc~\cite{li2025actloc}
      & SfM landmarks with colour, and the mapping camera poses
      & Rendered, a $6 \times 18$ grid of pitch and yaw at each position
      & hloc with SuperPoint
      & 0.1\,m\,/\,$1^\circ$
      & HM3D, 100 scenes; ScanNet~\cite{dai2017scannet}, 5 scenes \\ \addlinespace[3pt]
    Ours
      & Scan points and the depth of the mapping frames, without colour or descriptors
      & Real frames, four to eight per decision, with roll
      & SuperPoint matched to one reference image, EPnP~\cite{lepetit2009epnp} with RANSAC~\cite{fischler1981random}
      & 0.1\,m\,/\,$5^\circ$
      & SceneFun3D, real scans and videos \\
    \bottomrule
  \end{tabular}
  \begin{tablenotes}[para,flushleft]\footnotesize
  \item[]\hspace{-\labelsep}Map read by the planner: what the view planner uses to score a candidate. Our localizer still uses the descriptors of the reference image (Sec.~III-C3). The facts on prior work come from the papers, and for LWL also from its released code.
  \end{tablenotes}
  \end{threeparttable}
\end{table*}

\emph{Planner-side visibility.} The planner has no depth image of the
candidate, so it estimates from the map which landmarks $c$ sees. A
landmark is visible if it projects into the image and lies at most a
tolerance behind the nearest map point in its cell. In the dense tier, the
planner renders a z-buffer of the dense points with cells of 4\,px and a
tolerance of 5\,cm. A sparse point drawn as a single pixel would hide
nothing behind it. In the sparse tier, the planner therefore splats each
point as a square of the projected size of its voxel. The square is at most
seven cells wide, and the tolerance is $\delta$. The planner-visible
landmarks $\mathcal{V}(c) \subseteq \mathcal{L}_K(r(c))$ are the selected
landmarks that pass this test.

\emph{The angle planner.} Hanlon et al.~\cite{hanlon2024active} give no
thresholds for their angle heuristic. We implement two readings of it, and
both count the planner-visible landmarks that pass a test. In the first
reading, a landmark passes if one of its mapping observations lies within
$\alpha \in \{15, 30, 45, 60\}^\circ$ of the current viewing direction and
the ratio of the two distances lies in $[5/7, 7/5]$. This test follows
Davison and Murray~\cite{davison2002active}. In the second reading, the
current distance and the current angle to the optical axis must both lie
within their ranges over the mapping observations. The observations come
from every second mapping frame, and the depth of a frame decides within
5\,cm whether it observes a landmark. We choose the reading and $\alpha$ on
the validation set, separately for each landmark selection.

\emph{Features of the per-landmark planner.} For each landmark
$\ell \in \mathcal{V}(c)$, $\boldsymbol{\phi}_G$ holds 13 features. Five
features describe how $c$ sees the landmark. They are its depth, its angle
to the optical axis, its two normalized image coordinates and its distance
from the image centre. Three features relate this view to the reference
view. They are the distance of the landmark to the reference camera, the
angle between the viewing rays from $c$ and from $r(c)$, and the log ratio
of the two distances. Two features give the absolute cosine between the
surface normal and each of the two viewing rays. The last three features
count the other landmarks of $\mathcal{V}(c)$ within 32\,px in the image of
$c$, the map points within $2\delta$ of the landmark, and the landmarks in
$\mathcal{V}(c)$. In the dense tier, the radius of the second count is
2\,cm. In the sparse tier, $\delta$ is a further input, so one model serves
both voxel sizes. The normals there come from a principal component
analysis of the 16 nearest sparse points.

\emph{Classifier.} The per-landmark planner is the histogram
gradient-boosted tree classifier of scikit-learn~\cite{pedregosa2011sklearn}.
It uses 300 iterations, a learning rate of 0.05, at most 31 leaves, at least
200 samples per leaf and an L2 penalty of 1.0. It needs no feature
rescaling. We set these hyperparameters in a preliminary probe and did not
tune them afterwards.

\emph{Published baselines.} Our LWL~\cite{digiammarino2024lwl} bins the
landmarks on a $30 \times 30$ image grid. The per-bin means of their 2D,
depth or 3D coordinates feed an MLP with two hidden layers of 300 units and
dropout 0.5. The MLP is trained with a weighted binary cross-entropy on the
outcomes of our localizer, with the noise augmentation of the original. We
choose the input type, the learning rate and the epoch on the validation
set, over two landmark families. One family is the landmarks of the
localizer, and the other is all planner-visible map points. The chosen
setting is trained with three seeds, and the test uses the mean of their
probabilities. FIF~\cite{zhang2019fif,zhang2020fif} scores a view by the
Fisher information of the landmark bearings. We summarize this information
by its log-determinant, smallest eigenvalue or trace, with exact or
quadratic visibility, and choose one variant on the validation set.

\emph{Differences from prior protocols.} We follow the single-step view
selection of Hanlon et al.~\cite{hanlon2024active} and
LWL~\cite{digiammarino2024lwl}, with four differences
(Table~\ref{tab:settings}). First, their maps hold structure-from-motion
points triangulated from image features, so the density of the landmarks
encodes texture. ActLoc~\cite{li2025actloc} also gives its planner the
colour of each landmark. The map of our planner holds scan points and the
depth of the mapping frames, without descriptors or colour. Second, their
candidates are rendered views without roll. Our candidates are four to
eight real frames with roll. Real frames avoid the gap between rendered and
real images, but they give far fewer candidates per decision. Third, LWL
obtains its labels by localizing with SIFT~\cite{lowe2004sift} and
COLMAP~\cite{schoenberger2016sfm}, and Hanlon et al.\ localize with
hloc~\cite{sarlin2019hloc}. Our localizer matches
SuperPoint~\cite{detone2018superpoint} keypoints against the descriptors of
one reference image. Fourth, we use the single threshold of $5^\circ$ and
10\,cm. LWL reports four thresholds from 0.05\,m and $0.4^\circ$ upwards,
and Hanlon et al.\ report six from 0.025\,m and $1^\circ$ upwards. The
evaluation code of LWL divides the success rate by the success rate of the
oracle, and the table in the LWL paper appears to use this normalization.
We report both the raw success and the success relative to the oracle
(Table~IV). Absolute numbers are therefore not comparable across papers,
and only the orderings of planners can be compared.

\section{Experiment 2: further analyses of the sparse tier}
\label{sec:s-scope}
\begin{table}[t]
  \centering
  \caption{Experiment~2, sparse tier, test set: top-1 success (\%) under three endpoints}
  \label{tab:supp-rates}
  \footnotesize
  \setlength{\tabcolsep}{4pt}
  \begin{threeparttable}
  \begin{tabular}{@{}lcccccc@{}}
    \toprule
    & \multicolumn{3}{c}{Field landmarks} & \multicolumn{3}{c}{Harris landmarks} \\
    \cmidrule(lr){2-4}\cmidrule(l){5-7}
    Planner & $y$ & $\bar{y}$ & AUC & $y$ & $\bar{y}$ & AUC \\
    \midrule
    \meth{random} & 31.3 & 31.4 & 34.2 & 28.2 & 28.3 & 31.6 \\
    \meth{count} & 45.3 & 45.1 & 46.9 & 39.0 & 39.1 & 41.8 \\
    \meth{co-vis} & 44.8 & 44.7 & 48.1 & 40.4 & 40.5 & 44.7 \\
    \meth{Harris-sum} & 44.4 & 44.4 & 46.4 & 42.3 & 42.3 & 45.2 \\
    \meth{field-sum} & \textbf{49.0} & \textbf{48.8} & \textbf{50.9} & \textbf{43.6} & \textbf{43.7} & \textbf{46.4} \\
    \midrule
    \arm{G} & 50.2 & 50.3 & 53.0 & 44.6 & 44.7 & 48.4 \\
    \arm{G+H} & 51.1 & 51.0 & 53.7 & 46.5 & 46.4 & 50.0 \\
    \arm{G+C9} & 51.3 & 51.3 & 54.0 & 46.1 & 46.3 & 49.9 \\
    \arm{G+F} & \textbf{52.1} & \textbf{52.0} & \textbf{54.6} & \textbf{46.6} & \textbf{46.6} & \textbf{50.1} \\
    \arm{G+F+C9} & \textbf{52.3} & \textbf{52.2} & \textbf{54.9} & \textbf{46.7} & \textbf{46.8} & \textbf{50.4} \\
    \midrule
    \meth{oracle} & 77.9 & 69.1 & 68.0 & 74.0 & 63.7 & 63.7 \\
    \bottomrule
  \end{tabular}
  \begin{tablenotes}[para,flushleft]\footnotesize
  \item[]\hspace{-\labelsep}$y$: one localization attempt, the registered endpoint. $\bar{y}$: success rate over ten RANSAC runs on the same matches. Each run takes the matches in a new random order. AUC: success averaged over the ten runs and over the translation thresholds 5, 10, 25 and 50\,cm. Every threshold also requires a rotation error below $5^\circ$. $K$ and $\delta$ are averaged within each decision. Test set: 49 rooms, 4{,}385 decisions. The columns $y$ repeat Table~IV. \meth{angle} and the published baselines were evaluated with $y$ only (Tables~IV and VI).
  \end{tablenotes}
  \end{threeparttable}
\end{table}
\begin{table*}[t]
  \centering
  \caption{Experiment~2, sparse tier, test set: paired differences under three endpoints (percentage points)}
  \label{tab:supp}
  \footnotesize
  \setlength{\tabcolsep}{3pt}
  \begin{threeparttable}
  \begin{tabular}{@{}lccccc@{}}
    \toprule
    $\delta$, endpoint & H1: \arm{G+F}$-$\arm{G} & H2: \arm{G+F}$-$\arm{G+H} & H3: \arm{G+F}$-$\arm{G+C9}
      & \arm{G+F+C9}$-$\arm{G+C9} & \arm{G}$-$\meth{count} \\
    \midrule
    \multicolumn{6}{@{}l}{\emph{Field landmarks}} \\
    Merged, $y$ & \cisb{+1.8}{+1.3}{+2.4} & \cisb{+1.0}{+0.5}{+1.5} & \cisb{+0.8}{+0.3}{+1.4} & \cisb{+1.0}{+0.6}{+1.5} & \cis{+5.0}{+2.8}{+7.2} \\
    Merged, $\bar{y}$ & \cisb{+1.7}{+1.1}{+2.3} & \cisb{+0.9}{+0.5}{+1.4} & \cisb{+0.7}{+0.2}{+1.2} & \cisb{+0.9}{+0.5}{+1.4} & \cis{+5.2}{+3.0}{+7.6} \\
    Merged, AUC & \cisb{+1.6}{+1.1}{+2.2} & \cisb{+0.9}{+0.6}{+1.3} & \cisb{+0.6}{+0.2}{+1.1} & \cisb{+0.9}{+0.5}{+1.2} & \cis{+6.2}{+4.0}{+8.4} \\
    50\,mm, $y$ & \cisb{+2.6}{+1.9}{+3.3} & \cisb{+1.4}{+0.9}{+2.1} & \cisb{+1.2}{+0.6}{+1.9} & \cisb{+1.3}{+0.7}{+2.0} & \cis{+5.8}{+3.3}{+8.6} \\
    50\,mm, $\bar{y}$ & \cisb{+2.2}{+1.5}{+3.0} & \cisb{+1.3}{+0.8}{+1.9} & \cisb{+0.8}{+0.1}{+1.5} & \cisb{+1.1}{+0.6}{+1.7} & \cis{+6.0}{+3.4}{+8.8} \\
    100\,mm, $y$ & \cisb{+1.1}{+0.5}{+1.8} & \ci{+0.5}{-0.1}{+1.1} & \ci{+0.4}{-0.1}{+1.1} & \cisb{+0.7}{+0.2}{+1.3} & \cis{+4.1}{+2.1}{+6.0} \\
    100\,mm, $\bar{y}$ & \cisb{+1.1}{+0.6}{+1.7} & \cisb{+0.6}{+0.1}{+1.1} & \cisb{+0.6}{+0.1}{+1.0} & \cisb{+0.7}{+0.3}{+1.1} & \cis{+4.5}{+2.4}{+6.6} \\
    \midrule
    \multicolumn{6}{@{}l}{\emph{Harris landmarks}} \\
    Merged, $y$ & \cisb{+1.9}{+1.1}{+2.9} & \ci{+0.1}{-0.5}{+0.7} & \cisb{+0.4}{+0.0}{+0.9} & \cisb{+0.6}{+0.2}{+1.0} & \cis{+5.6}{+3.8}{+7.4} \\
    Merged, $\bar{y}$ & \cisb{+1.8}{+1.1}{+2.7} & \ci{+0.2}{-0.4}{+0.7} & \ci{+0.3}{-0.1}{+0.8} & \cisb{+0.5}{+0.1}{+0.9} & \cis{+5.7}{+3.9}{+7.4} \\
    Merged, AUC & \cisb{+1.7}{+1.1}{+2.4} & \ci{+0.1}{-0.3}{+0.5} & \ci{+0.3}{-0.1}{+0.6} & \cisb{+0.5}{+0.2}{+0.8} & \cis{+6.6}{+4.9}{+8.3} \\
    50\,mm, $y$ & \cisb{+2.5}{+1.4}{+3.7} & \ci{+0.3}{-0.3}{+1.2} & \ci{+0.5}{-0.1}{+1.2} & \cisb{+0.8}{+0.2}{+1.4} & \cis{+5.8}{+4.0}{+7.6} \\
    50\,mm, $\bar{y}$ & \cisb{+2.2}{+1.3}{+3.3} & \ci{+0.4}{-0.2}{+1.2} & \ci{+0.2}{-0.3}{+0.9} & \cisb{+0.5}{+0.0}{+1.1} & \cis{+6.3}{+4.5}{+8.0} \\
    100\,mm, $y$ & \cisb{+1.4}{+0.6}{+2.4} & \ci{-0.2}{-0.8}{+0.5} & \ci{+0.3}{-0.1}{+0.8} & \cisb{+0.4}{+0.0}{+0.8} & \cis{+5.4}{+3.4}{+7.5} \\
    100\,mm, $\bar{y}$ & \cisb{+1.5}{+0.8}{+2.3} & \ci{-0.1}{-0.6}{+0.4} & \ci{+0.4}{-0.0}{+0.8} & \cisb{+0.4}{+0.1}{+0.8} & \cis{+5.0}{+3.2}{+7.0} \\
    \bottomrule
  \end{tabular}
  \begin{tablenotes}[para,flushleft]\footnotesize
  \item[]\hspace{-\labelsep}Endpoints as in Table~\ref{tab:supp-rates}. Merged: $\delta = 50$ and 100\,mm averaged within each decision. The 95\% intervals come from a room-clustered bootstrap with 20{,}000 resamples. Field landmarks, merged $\delta$ and $y$ form the registered main decision, and field landmarks at 100\,mm with $y$ the registered secondary hypothesis. All other rows are descriptive. The rows with $y$ repeat rows (a)--(d) of Table~V. AUC is reported for merged $\delta$ only.
  \item[$*$] The interval excludes 0. A lower end printed as $+0.0$ is positive before rounding, and one printed as $-0.0$ is negative.
  \end{tablenotes}
  \end{threeparttable}
\end{table*}

The first four analyses below use the test-set run in the sparse tier
(Sec.~V-B2, Table~V). None of them changes the registered decision.

\emph{Voxel size.} The registered secondary hypothesis repeats H1--H3 at
$\delta = 100$\,mm (Table~V, row~(b)). Only H1 holds there, with
\cis{+1.1}{+0.5}{+1.8}. H2 is \ci{+0.5}{-0.1}{+1.1}, and H3 is
\ci{+0.4}{-0.1}{+1.1}. On the validation set, H1 and H2 had held at
100\,mm only, with \cis{+1.6}{+0.4}{+3.1} and \cis{+0.7}{+0.0}{+1.5}. We
registered the secondary hypothesis for this reason. This pattern did not
replicate. On the test set, the effects are larger at 50\,mm (row~(c)).

\emph{Landmark selection.} With Harris landmarks, H1 still holds
(\cis{+1.9}{+1.1}{+2.9}), but H2 does not. \arm{G+F}$-$\arm{G+H} is
\ci{+0.1}{-0.5}{+0.7}. With these landmarks, the Harris input alone adds
as much as the field, and \arm{G+H}$-$\arm{G} is \cis{+1.9}{+1.4}{+2.4}.
The advantage over Harris-3D is therefore detectable only when the field
also selects the landmarks. Our claim over Harris-3D thus concerns the
field used both to choose the landmarks and to plan, which is the sparse
localization system of \fieldname{}. It does not concern the field as a
planner input alone.

\emph{Scalar planners.} With field landmarks, \meth{field-sum} beats both
\meth{count} and \meth{Harris-sum}, by \cis{+3.8}{+1.6}{+5.7} and
\cis{+4.6}{+2.5}{+6.9}. On the validation set, its margin over
\meth{Harris-sum} was \cis{+3.5}{+0.4}{+6.2}. This margin over the Harris
counterpart is larger than that of any learned arm. With Harris
landmarks, the lead of \meth{field-sum} over \meth{Harris-sum} shrinks to
\cis{+1.4}{+0.2}{+2.7}. This suggests that most of the scalar advantage
with field landmarks comes from summing the same field that chose the
landmarks. We therefore do not read it as independent predictive power of
the field. With field landmarks, \meth{angle} beats \meth{count}
(\cis{+1.5}{+0.5}{+2.4}), and \meth{co-vis} does not differ from
\meth{count} (\ci{-0.5}{-4.0}{+2.6}). With Harris landmarks, \meth{angle}
is level with \meth{count} (\ci{-0.8}{-1.6}{+0.1}).

\emph{Endpoint and noise.} We replaced $y$ by the success rate over ten
reshuffled RANSAC runs on the same matches, and by an area under the recall
curve over four translation thresholds. All three main comparisons keep
intervals above zero (Tables~\ref{tab:supp-rates} and \ref{tab:supp}). The
outcomes of the protocol are noisy. The same candidate has the same outcome
at all three values of $K$ in only 67.0\% of cases, and in 61.8\% on the
validation set. Two known sources may contribute. First, the adaptive
stopping rule of RANSAC relies on an approximation of the sampling
probability. This approximation leads to too few samples, especially when
inliers are few~\cite{schoenberger2025ransacstop}. Our localizer uses this
stopping rule in OpenCV, with a confidence of 0.999 and at most 2000
iterations. Second, the choice of the reference algorithm for pseudo ground
truth can change the ranking of localization methods~\cite{brachmann2021pseudogt}.
We have not measured the accuracy of the frame poses of SceneFun3D against
our 10\,cm threshold. The localizer was fixed by the registration, and we
did not change it.

\emph{A different PnP solver.} An earlier probe used a different protocol,
with scalar utilities only and other rooms. In this probe, we replaced the
EPnP-RANSAC of OpenCV with a batched P3P solver. With this solver,
\meth{random} succeeded far more often, on one room set in 62\% of the
decisions instead of 39\%. The advantage of field landmarks with
field-weighted planning over the Harris-3D counterpart did not persist on
any of the three room sets. On the room set above, the differences among
planners vanished. We did not repeat this probe under the present protocol,
so our results hold for the registered localizer only.

\section{Experiment 2: the dense tier}
\label{sec:s-dense}

\begin{table}[t]
  \centering
  \caption{Experiment~2, dense tier: top-1 success (\%)}
  \label{tab:dense}
  \footnotesize
  \begin{threeparttable}
  \begin{tabular}{@{}lcc@{}}
    \toprule
    Planner & Validation & Test \\
    \midrule
    \meth{random}      & 49.9 & 43.5 \\
    \meth{count}       & 64.8 & 60.4 \\
    \meth{co-vis}      & 65.1 & 59.5 \\
    FIF                & 59.3 & 55.1 \\
    \meth{angle}       & 66.8 & 61.3 \\
    \meth{Harris-sum}  & 66.2 & 61.9 \\
    \meth{field-sum}   & 66.4 & \textbf{64.2} \\
    \midrule
    \arm{G}            & 69.3 & 60.7 \\
    \arm{G+H}          & 70.8 & 63.8 \\
    \arm{G+C9}         & 71.3 & 64.7 \\
    \arm{G+F}          & \textbf{71.6} & 64.1 \\
    \arm{G+F+C9}       & \textbf{72.0} & \textbf{65.4} \\
    \midrule
    \meth{oracle}      & 85.3 & 87.4 \\
    \bottomrule
  \end{tabular}
  \begin{tablenotes}[para,flushleft]\footnotesize
  \item[]\hspace{-\labelsep}Harris landmarks, endpoint $y$, $K$ averaged within each decision. Validation: 12 rooms, 1{,}913 decisions. Test: 49 rooms, 4{,}412 decisions.
  \end{tablenotes}
  \end{threeparttable}
\end{table}

Table~\ref{tab:dense} lists the top-1 success of all planners in the dense
tier. Rows (f) and (g) of Table~V give the paired differences, and
Sec.~V-B3 discusses them. On the test set, H1 holds
(\cis{+3.5}{+2.6}{+4.4}). H2 does not hold (\ci{+0.3}{-0.4}{+1.1}), and
neither does H3 (\ci{-0.6}{-1.2}{+0.0}). Added to the curvature arm, the
field still raises the success by \cis{+0.8}{+0.4}{+1.2}. In place of the
curvature input, it does no better.

Two observations on the test set differ from the validation set. First,
\arm{G} no longer beats \meth{count}. Its margin is \ci{+0.3}{-1.9}{+3.0}
on the test set and \cis{+4.4}{+1.1}{+8.4} on the validation set. Second,
\meth{field-sum} succeeds in 64.2\% of the test decisions, close to the
learned arms that use the field (64.1\% and 65.4\%). On the validation
set, it was well below them (66.4\% against 71.6\% and 72.0\%). In the
test rooms, a
random candidate also succeeds less often than in the validation rooms, in
43.5\% of the decisions against 49.9\% (Table~\ref{tab:dense}). This
difference may be related to the two observations, but we have not tested
it.

\section{Experiment 2: why the gain is small}
\label{sec:s-small}

\emph{Diagnosis (validation set).} We trained \arm{G} with an ideal static
field $\bar p$. For each landmark, $\bar p$ is the fraction of the other
decisions in the same crop in which the landmark was matched correctly
when the planner found it visible. The fraction is smoothed towards the
mean of the crop. It approximates the most that a field without
dependence on the viewpoint could supply. The estimate is optimistic,
because it uses outcomes from the same crop. In the sparse tier with field
landmarks, $\bar p$ adds only \ci{+3.0}{-0.2}{+6.3} points over \arm{G},
with 57.8\% against 54.8\%. The field of \fieldname{} obtains 44\% of this
gain. In the dense tier, $\bar p$ adds \cis{+3.0}{+0.8}{+5.7}, and the
dense field obtains 76\% of it. Per landmark, $\bar p$ separates correct
from incorrect matches far better than the field. In the sparse tier with
field landmarks, the AUC is 0.889 against 0.542. However, the utility of a
candidate sums over tens of landmarks, and the geometric features already
explain most differences between candidates. The field predicts correct
matching only weakly, and it predicts even less whether a correct match
then becomes a RANSAC inlier (AUC 0.519). This second step depends on
geometry, and the depth of the landmark reaches an AUC of 0.667 there. Even
a planner that knows the true number of correct matches of each candidate
reaches only 58.8\%, against 80.4\% for the oracle. This suggests that much
of the remaining gap is outcome noise, which a planner cannot predict from
the map.

\emph{Other uses of the field.} During development, we also tried four
other ways to use the field in the sparse tier. Two utilities summarized the
field rendered over the whole candidate view, one by its coverage of the
view and one by Fisher information. A third utility aggregated the
per-landmark probabilities of the planner by Fisher information instead of
summing them. A fourth added statistics of the rendered field around each
landmark to the per-landmark features. None of them beat the per-landmark
planner with the field, and the two rendered utilities fell below
\meth{count}. We therefore use the field only as a per-landmark input.
These attempts were exploratory, and we report them only to explain this
choice.

\section{Experiment 2: transfer to HM3D}
\label{sec:s-hm3d}

As a descriptive check on a second dataset, we rendered handheld-like
trajectories in two scenes from HM3D~\cite{ramakrishnan2021hm3d}, using the
meshes distributed with ActLoc~\cite{li2025actloc}. The speed, the angular
speed, the camera height and the pitch of the trajectories follow the
statistics of SceneFun3D, and each room has its own video. We ran the
sparse-tier pipeline unchanged, with the planners trained on SceneFun3D.
As on SceneFun3D, the two arms with the field succeed most often.
\arm{G+F+C9} succeeds in 68.5\% of the decisions and \arm{G+F} in
68.2\%. They are followed by \arm{G+H} and \arm{G+C9} (67.3\%), \arm{G}
(65.5\%) and \meth{count} (61.4\%). \arm{G+F}$-$\arm{G} is \cis{+2.8}{+1.2}{+4.5}. The differences
to \arm{G+H} and \arm{G+C9} are small and positive, and their intervals
include zero (\ci{+0.9}{-0.3}{+2.5} and \ci{+0.9}{-0.4}{+2.7}). In
addition, \arm{G+F+C9}$-$\arm{G+C9} is \cis{+1.2}{+0.0}{+2.6}. The
evidence is thin. It
comes from two scenes, 28 crops with decisions and 127 decisions, and the
intervals are clustered by crop rather than by room.

We departed once from the registered design of this check. A geometry
check, registered before any localization run, allowed at most 5\% of the
queries to fail a visibility test. A query fails it when the candidate
truly sees less than a tenth of the landmarks that the planner finds
visible. Planner-side visibility uses only the points of the crop, so walls
outside the crop do not occlude. In one room video, the planner therefore
treated the crop as visible through walls from neighbouring rooms, and
7.8\% of all queries failed the test. The registered rule was to stop. We
instead dropped the crops of that video, after this check and before seeing
any localization outcome. The HM3D result is therefore descriptive only.

\bibliographystyle{IEEEtran}
\bibliography{refs}